\documentclass[10pt,twocolumn,letterpaper]{article}

\usepackage{cvpr}              

\usepackage{tocloft}

\definecolor{cvprblue}{rgb}{0.21,0.49,0.74}
\usepackage[pagebackref,breaklinks,colorlinks,allcolors=cvprblue]{hyperref}
\usepackage{multirow} 

\def\paperID{804} 
\def\confName{CVPR}
\def\confYear{2026}

\title{PerformRecast: Expression and Head Pose Disentanglement \\ for Portrait Video Editing\vspace{-5mm}}

\author{Jiadong Liang\textsuperscript{*} \quad
    Bojun Xiong\textsuperscript* \quad
    Jie Tian \quad
    Hua Li \quad
    Xiao Long \quad
    Yong Zheng \quad
    Huan Fu\textsuperscript{\dag} 
    \\[0.5em]
    HUJING Digital Media \& Entertainment Group
    \\[0.5em]
    {\small \textsuperscript{*}Equal contribution \quad \textsuperscript{\dag}Corresponding author}
}

\makeatletter
\let\@oldmaketitle\@maketitle%
\renewcommand{\@maketitle}{\@oldmaketitle%
 \centering
 \vspace{-5mm}
    \includegraphics[width=0.95\textwidth]{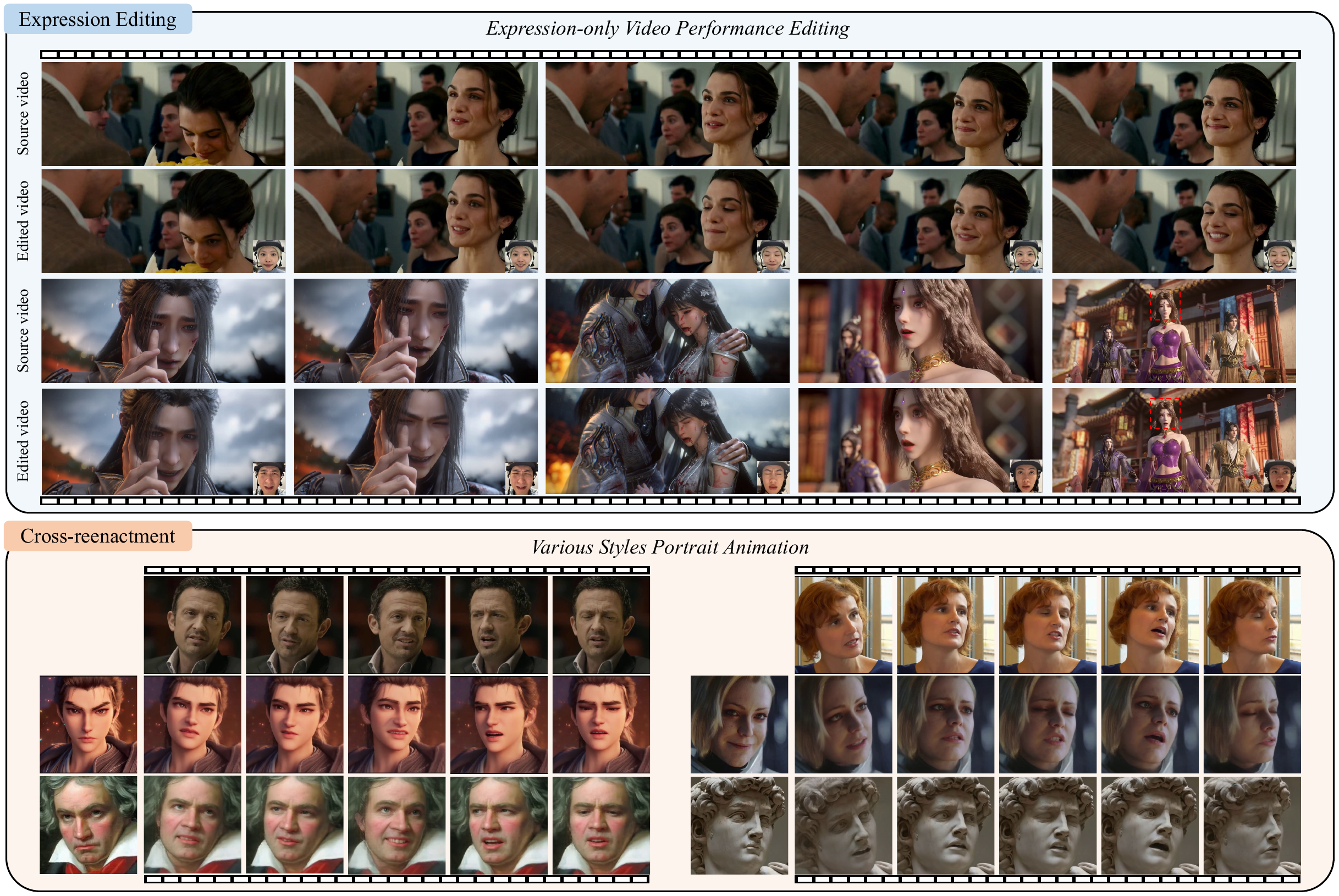} 
    \vspace{-3mm}
     \captionof{figure}{Our proposed PerformRecast is capable of editing the facial expression of a source portrait video as well as animating a static portrait image according to a driving video. The top part of this figure shows the expression editing results of a movie clip and a 3D animation. The generated results exhibit high fidelity to the driving video, facilitating the production processes in film and animation industries. For the shot with multiple characters, we can select the specific person whose facial expression we want to edit, which is indicated by a red dashed box. The bottom-right insets on the top part show the driving frames. Please zoom in for better inspection.}
    \label{fig:teaser}
    \bigskip}
\makeatother

\begin{document}
\maketitle
\begin{abstract}
This paper primarily investigates the task of expression-only portrait video performance editing based on a driving video, which plays a crucial role in animation and film industries. Most existing research mainly focuses on portrait animation, which aims to animate a static portrait image according to the facial motion from the driving video. As a consequence, it remains challenging for them to disentangle the facial expression from head pose rotation and thus lack the ability to edit facial expression independently. In this paper, we propose PerformRecast, a versatile expression-only video editing method which is dedicated to recast the performance in existing film and animation. The key insight of our method comes from the characteristics of 3D Morphable Face Model (3DMM), which models the face identity, facial expression and head pose of 3D face mesh with separate parameters. Therefore, we improve the keypoints transformation formula in previous methods to make it more consistent with 3DMM model, which achieves a better disentanglement and provides users with much more fine-grained control. Furthermore, to avoid the misalignment around the boundary of face in generated results, we decouple the facial and non-facial regions of input portrait images and pre-train a teacher model to provide separate supervision for them. Extensive experiments show that our method produces high-quality results which are more faithful to the driving video, outperforming existing methods in both controllability and efficiency. Our code, data and trained models are available at \url{https://youku-aigc.github.io/PerformRecast}.
\end{abstract}    
\vspace{-5mm}
\section{Introduction}
\label{sec:intro}

Character animation with vivid facial expression is of vital importance in the film and computer animation industries~\cite{naruniec2020high}. However, it is quite challenging for artists to create lifelike expression on 3D face models, and for film actors to consistently perform satisfying facial expressions, whether in a single take or with minimal cost. Therefore, developing an automatic algorithm on expression-only portrait video performance editing would be highly meaningful and important in the field of computer animation.

With the rapid development of deep generative models, such as GANs~\cite{goodfellow2014generative, karras2020analyzing} and Diffusion Models~\cite{rombach2022high, ho2020denoising, song2020denoising}, most existing studies mainly focus on animating a static portrait image following the facial motion of a driving video, also termed portrait animation, which is a little different from our target. Given a video (e.g., a movie clip), our goal is to edit or enhance only the facial expression based on a real actor's time-varying facial expression, while strictly preserving all other factors, including the original face ID, head pose, camera motion and background. Any other change outside facial expression is considered as a failure. Diffusion-based portrait animation methods are typically built upon pre-trained image diffusion models~\cite{rombach2022high, SD1-5} or video diffusion models~\cite{blattmann2023stable, hong2022cogvideo, yang2024cogvideox, wan2025wan} via attaching additional modules. But they struggle to fully disentangle the facial expression from head pose rotation, making it difficult to accurately perform our expression editing task. 

On the other hand, recent GAN-based portrait animation methods typically construct a 3D feature volume given source and driving portrait images~\cite{wang2021facevid2vid, Drobyshev22MP, drobyshev2024emoportraits, guo2024liveportrait}, which is further sent into the generator to produce the final image. These methods employ motion encoder to extract corresponding features~\cite{Drobyshev22MP, drobyshev2024emoportraits} or implicit keypoints~\cite{wang2021facevid2vid, guo2024liveportrait} to guide the computation of feature volume. However, since the extracted features are obtained directly from raw images, they still have difficulties in disentangling face identity, facial expression and head pose. Moreover, the implicit keypoints lack explicit physical meaning and direct supervision, thereby compromising the precision of facial motion control and leading to suboptimal results.

In this paper, we propose PerformRecast, a well-designed and effective GAN-based method tailored to our expression-only video performance editing task. Traditional 3DMM~\cite{3dmm1999} represents face identity, facial expression, and head pose with separate parameters, naturally disentangling these factors. Therefore, we improve the keypoints transformation used in previous method to make it more consistent with the forward process of FLAME~\cite{FLAME2017}, a representative 3D Morphable Face Model. Specifically, we employ a 3D face tracking method~\cite{giebenhain2025pixel3dmm} to extract temporally continuous FLAME~\cite{FLAME2017} parameters from input portrait videos, and select explicit 3D keypoints on face mesh vertices to supervise the motion extractor in our model. To further avoid the misalignment around the boundary of the face in the generative results, we introduce a boundary alignment module which segments each frame into facial and non-facial regions. An additional teacher model is pre-trained to provide the training loss for the facial region, enabling separate supervision for both regions. As a result, our method can not only edit and enhance facial expressions in existing videos due to its disentanglement from head pose, but also outperform many previous methods on the traditional portrait animation task. In addition, to better evaluate expression editing performance, we construct a benchmark using digital humans from MetaHuman~\cite{metahuman}. For each digital human, we render multiple portrait videos with the same head pose rotation but different facial expressions. In summary, the contributions of our paper are fourfold:
\begin{itemize}
\item We modify the keypoints transformation formula and utilize explicit 3D keypoints on FLAME face mesh to directly supervise the motion extractor.
\item We adopt a boundary alignment module to alleviate misalignment between the facial and non-facial regions.
\item We present a benchmark, tailored for the assessment of portrait video expression editing, which will be publicly available to advance the evaluation of future research.
\item We propose PerformRecast, a versatile and effective method which is expert at both expression-only video performance editing and portrait animation tasks. Qualitative and quantitative experiments have been conducted to verify the superiority of our method over other existing approaches in controllability and quality.
\end{itemize}
\vspace{-2mm}
\section{Related Work}
\label{sec:related_work}
In this section, we mainly summarize different types of portrait animation methods, for they are quite similar to our expression editing task from a methodological perspective.

\subsection{Non-Diffusion-based Portrait Animation}
\label{sec:non-diffusion}

Early 3D face model-based methods~\cite{ma2019real, thies2016face2face} reconstruct high-quality geometry and appearance for rendering. With the development of 3D Morphable Face Models~\cite{3dmm1999, FLAME2017} and neural rendering~\cite{thies2019deferred, mildenhall2020nerf, kerbl3Dgaussians}, methods such as Portrait4D~\cite{deng2024portrait4d, deng2024portrait4dv2} and GAGAvatar~\cite{chu2024gagavatar} adopt triplane or 3D Gaussian representations for animatable head synthesis. However, pure 3D-based representations often struggle to capture fine details, leading to blurry results.

A great deal of other methods are built upon the
Generative Adversarial Networks~\cite{goodfellow2014generative, karras2019style, karras2020analyzing}, which provide stronger image synthesis capabilities. Early approaches directly decode latent appearance and motion features~\cite{zakharov2019few,burkov2020neural}, while later works focus on disentangling identity and motion via specialized designs~\cite{zeng2020realistic, wang2022pdfgc, tan2025fixtalk, tan2024edtalk}. Nevertheless, these methods rely heavily on complex loss functions and still face challenges in achieving complete disentanglement.

More recent portrait animation models are predominantly warping-based~\cite{siarohin2019animating, siarohin2019first, siarohin2021motion} methods, which estimate motion fields using learned landmarks/keypoints and warp source features~\cite{wiles2018x2face, zakharov2020fast,Doukas_2021_ICCV, yang2022face2face, hong2022depth, ren2021pirenderer}. Representative works such as LIA-X~\cite{wang2022latent,wang2025lia}, EMOPortrait~\cite{Drobyshev22MP,drobyshev2024emoportraits}, Face Vid2vid~\cite{wang2021facevid2vid} and LivePortrait~\cite{guo2024liveportrait} improve motion modeling and controllability. However, their implicit motion representations lack explicit physical meaning and supervision, which restricts the control flexibility and accuracy.

\vspace{-1mm}
\subsection{Diffusion-based Portrait Animation}
\vspace{-1mm}
\label{sec:diffusion}

With the rapid development of diffusion models~\cite{ho2020denoising, song2020denoising}, recent works leverage large-scale pre-trained image and video diffusion models~\cite{rombach2022high, SD1-5, blattmann2023stable, hong2022cogvideo, yang2024cogvideox, wan2025wan} for portrait animation. These methods typically incorporate several plug-and-play modules to capture identity, background content as well as facial motion and maintain the cross-frame coherence. 

Some representative approaches include Follow-Your-Emoji~\cite{ma2024follow}, X-NeMo~\cite{zhao2025x}, Wan-Animate~\cite{cheng2025wan}, VACE~\cite{vace}, Hunyuan-Portrait~\cite{xu2025hunyuanportrait}, AniPortrait~\cite{wei2024aniportrait}, SkyReels-A1~\cite{qiu2025skyreels}, and AvatarArtist~\cite{liu2025avatarartist}. They adopt diverse motion representations such as landmarks, latent motion codes, 3D priors, or implicit features. However, despite the various motion representations they used, they can hardly disentangle the face identity, facial expression and head pose. What’s more, they struggle to guarantee temporal consistency and require much more inference time.
\vspace{-2mm}
\section{Method}
\label{sec:method}

\begin{figure*}[t!]
  \centering
  \includegraphics[width=\textwidth]{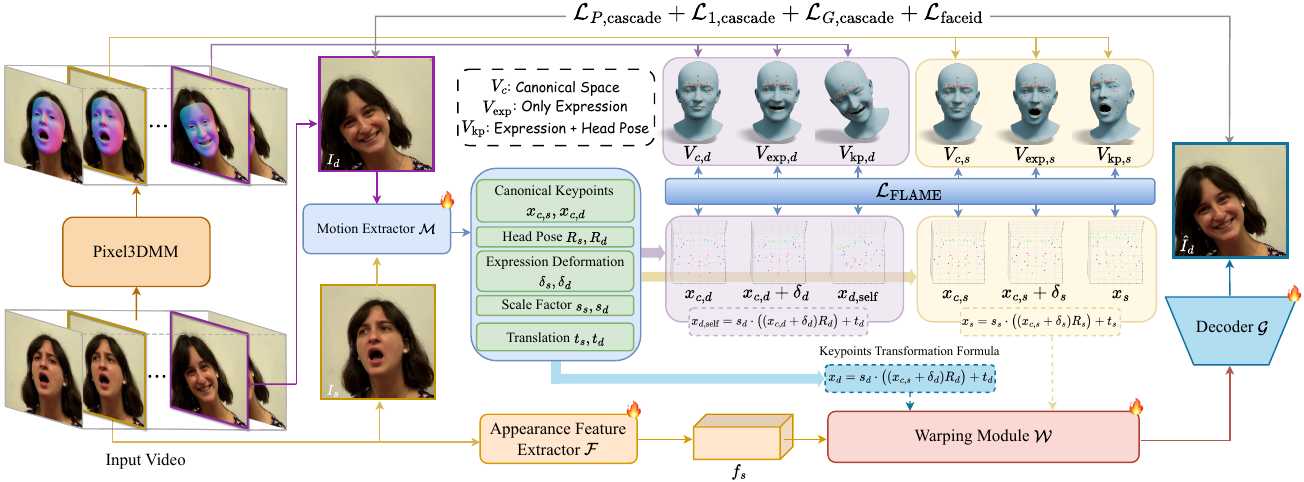}
  \vspace{-8mm}
  \caption{An overview of our PerformRecast framework. The motion extractor extracts canonical keypoints, head pose, expression deformation, scale factor and translation from source and driving frames. The facial keypoints are obtained via our improved keypoints transformation formula and compared with the tracking results from Pixel3DMM~\cite{giebenhain2025pixel3dmm} to calculate the FLAME loss. Finally, the appearance feature volume, source and driving keypoints are sent to the warping module, followed by the decoder to reconstruct the driving frame.}
  \label{fig:method}
\vspace{-5mm}
\end{figure*}

\vspace{-1mm}
In this section, we provide a detailed explanation of our method, PerformRecast. Our method is built upon LivePortrait~\cite{guo2024liveportrait}, a typical warping-based portrait animation model. The overall training pipeline of our method is shown in~\cref{fig:method}. We change the original keypoints transformation formula used in LivePortrait to make it more consistent with 3D face parametric model which is naturally capable of disentangling the face identity, facial expression and head pose. Then, we present our boundary alignment module which alleviates the misalignment between the facial and non-facial regions in generated images. Finally, we elaborate on the inference process of our method.

\subsection{Preliminary}
We begin with a brief review of LivePortrait~\cite{guo2024liveportrait} and 3DMM-based face tracking~\cite{giebenhain2025pixel3dmm}, upon which our method builds. LivePortrait~\cite{guo2024liveportrait} utilizes the self-reenactment training pipeline, which takes source and driving frames within the same subject and video. We denote the source and driving frames as $I_s \in \mathbb{R}^{3\times H\times W}$ and $I_d\in \mathbb{R}^{3\times H\times W}$. The model is learned to reconstruct the driving frame $I_d$, and the synthesized frame is denoted as $\hat{I}_d$. The original framework consists of an appearance feature extractor $\mathcal{F}$, a motion extractor $\mathcal{M}$, a warping field estimator $\mathcal{W}$ and a SPADE decoder~\cite{park2019semantic} based generator $\mathcal{G}$. $\mathcal{F}$ maps the source portrait image $I_s$ to a 3D appearance feature volume $f_s$. The motion extractor $\mathcal{M}$, with ConvNext-V2-Tiny~\cite{woo2023convnext} backbone, directly predicts the canonical keypoints $x_c \in \mathbb{R}^{K \times 3}$, head pose $R\in\mathbb{R}^{3 \times 3}$, expression deformation $\delta \in \mathbb{R}^{K \times 3}$, scale factor $s \in \mathbb{R}^3$ and translation $t \in \mathbb{R}^3$ from both source and driving frames. $K$ represents the number of implicit keypoints. Then, the source 3D keypoints $x_s$ and the driving 3D keypoints $x_d$ are transformed as follows:
\vspace{-3mm}
\begin{equation}
    \left\{
        \begin{array}{lr}
        x_{s} = s_s \cdot (x_{c,s} R_{s} + \delta_{s}) + t_{s}, &  \\
        x_{d} = s_d \cdot (x_{c,s} R_{d} + \delta_{d}) + t_{d}, &
        \end{array}
    \right.
    \label{equ:LP_ori}
\vspace{-1mm}
\end{equation}
where the subscripts $s$ and $d$ denote the source and driving, respectively. Next, $\mathcal{W}$ generates the warping field using the implicit keypoints $x_s$ and $x_d$, and employs this flow field to warp the source feature volume $f_s$. Finally, the warped features pass through the generator $\mathcal{G}$, which translates them into image space and generates a target image.

What's more, We adopt a recently-proposed 3D face tracking method, Pixel3DMM~\cite{giebenhain2025pixel3dmm} to predict FLAME parameters of each frame from input portrait videos. The FLAME parameters include face identity $\beta \in \mathbb{R}^{300}$, expression $\psi \in \mathbb{R}^{100}$, head pose $\theta \in \mathbb{R}^{3 * 4 + 3 = 15}$ and other camera parameters. The head pose $\theta$ contains four 3D rotation vectors for four joints: $\theta_\text{neck}, \theta_\text{jaw}, \theta_\text{left-eyeball}$, $\theta_\text{right-eyeball}$ and one global rotation $\theta_\text{head}$ in axis-angle. We describe the details of Pixel3DMM~\cite{giebenhain2025pixel3dmm} in the supplementary material.

\vspace{-1mm}
\subsection{FLAME-based Keypoints Transformation}
\vspace{-1mm}
Similar to LivePortrait~\cite{guo2024liveportrait}, our method also utilize the self-reenactment training pipeline, which reenacts both the facial expression and head pose to reconstruct the driving frame $I_d$ from source frame $I_s$. We argue that the implicit keypoints transformation used in LivePortrait can not fully disentangle the face identity, facial expression and head pose. The canonical keypoints $x_c$ are first multiplied by the head pose $R$ and then added with the expression deformation $\delta$. As a consequence, the learned expression deformation would contain some residual head pose information. On the contrary, the traditional 3D Morphable Face Model (3DMM)~\cite{3dmm1999}, such as FLAME~\cite{FLAME2017}, uses separate parameters to represent identity, expression, and head pose, achieving a natural disentanglement. The transformation process of FLAME~\cite{FLAME2017} is defined as:
\vspace{-2mm}
\setlength{\belowdisplayskip}{5pt}
\begin{equation}
\resizebox{0.9\linewidth}{!}{$
\begin{split}
& M(\beta, \theta, \psi) = W(T_P(\beta, \theta, \psi), \mathbf{J}(\beta), \theta, \mathcal{W}), \\
& T_P(\beta, \theta, \psi) = \mathbf{T} + B_S(\beta;\mathcal{S}) + B_P(\theta; \mathcal{P}) + B_E(\psi; \mathcal{E}), \end{split}
$}
\label{equ:flame}
\end{equation}
This function takes different coefficients to describe shape $\beta$, head pose $\theta$, expression $\psi$ and returns $N$ vertices. \cref{equ:flame} illustrates that the template mesh $\mathbf{T}$, in the zero pose and expression, is first added with identity related shape variation $B_S(\beta; \mathcal{S})$ and expression blendshapes $B_E(\psi;\mathcal{E})$, then multiplied by head pose rotation $\theta$ (the pose blendshapes term $B_P(\theta; \mathcal{P})$ is not necessary in our task). 

Therefore, we modify the keypoints transformation formula used in LivePortrait to the scale orthographic projection~\cite{guo2024liveportrait}, which is formulated as:
\vspace{-2mm}
\setlength{\belowdisplayskip}{8pt}
\begin{equation}
    \left\{
        \begin{array}{lr}
        x_{s} = s_s \cdot \big((x_{c,s} +\delta_{s}) R_s \big) + t_{s}, &  \\
        x_{d} = s_d \cdot \big((x_{c,s} + \delta_{d}) R_d \big) + t_{d}, &
        \end{array}
    \right.
    \label{equ:keypoint}
\end{equation}
The scale orthographic projection is more similar with FLAME model, whose canonical keypoints are first added with the expression deformation $\delta$ and then multiplied by the head pose $R$, effectively avoiding information leakage between head pose and facial expression. 

What's more, instead of treating $x_s$ and $x_d$ as implicit keypoints, we select several explicit keypoints from vertices of FLAME face mesh tracked by Pixel3DMM~\cite{giebenhain2025pixel3dmm} to directly supervise the keypoints transformation process in \cref{equ:keypoint}. Specifically, we derive three sets of explicit keypoints from the reconstructed FLAME model of source and driving frame. The first set of explicit keypoints $V_{c,i}$ is obtained from the canonical FLAME face mesh $T_{c,i}$, which encodes only shape blendshapes by setting both head pose  $\theta$ expression $\psi$ to zero. The second set $V_{\text{exp},i}$, which is used to supervise the expression deformation, is extracted from the FLAME face mesh $T_{\text{exp},i}$. $T_{\text{exp},i}$ adds expression blendshapes and three joint rotations $\theta_\text{jaw}, \theta_\text{left-eyeball}, \theta_\text{right-eyeball}$ to $T_c$ and keeps neck pose $\theta_\text{neck}$ as well as global head pose $\theta_\text{head}$ to zero, for facial expression often contains the orientation of eyeballs and the movement of jaw, while excluding the rotation of neck in most scenarios. The third set, $V_{\text{kp},i}$ is derived from $T_{\text{kp},i}$, where all FLAME parameters are enabled. The subscripts $i \in \{s,d\}$ denotes the source and driving frames, respectively. We then introduce the FLAME loss which utilizes these three sets of explicit keypoints to directly supervise the motion extractor. The FLAME loss is formulated as follows:
\vspace{-1mm}
\setlength{\belowdisplayskip}{5pt}
\begin{align}
\mathcal{L}_{\text{FLAME}} = {} &
\text{Wing}(x_{c,s}, V_{c,s}) + \text{Wing}(x_{c,d}, V_{c,d}) \notag \\
+ & ~ \text{Wing}(x_{c,s} + \delta, V_{\text{exp}, s}) + \text{Wing}(x_{c,d} + \delta, V_{\text{exp}, d}) 
\notag \\
+ & ~ \text{Wing}(x_s, V_{\text{kp}, s}) + \text{Wing}(x_{d, \text{self}}, V_{\text{kp}, d}),
\vspace{-2mm}
\end{align}
where $x_{d, \text{self}} = s_d \cdot \big((x_{c,d} + \delta_{d}) R_d \big) + t_{d}$, which is additionally computed to accelerate training and Wing loss is adopted following~\cite{feng2018wing}. The FLAME loss provides a much stronger supervision to motion extractor, enabling more accurate learning of keypoint transformations. It also prevents the model from learning overly flexible expressions $\delta$ as mentioned in \cite{guo2024liveportrait}. Therefore, we discard the implicit keypoints equivariance loss, keypoint prior loss, deformation prior loss and head pose loss used in previous methods~\cite{wang2021facevid2vid, guo2024liveportrait} for head pose $R$ can be effectively learned via self supervised learning. The overall training loss of our model is similar to that of previous portrait animation methods, which is formulated as:
\vspace{-2mm}
\begin{equation}
\resizebox{0.9\linewidth}{!}{$
    \mathcal{L}_\text{animate} = \mathcal{L}_\text{FLAME} + \mathcal{L}_{P,\text{cascade}} + \mathcal{L}_{1,\text{cascade}} +\mathcal{L}_{G, \text{cascade}} + \mathcal{L}_\text{faceid},$}
\label{equ:animate}
\end{equation}
where the cascaded perceptual loss $\mathcal{L}_{P,\text{cascade}}$, the cascaded $L_1$ loss $\mathcal{L}_{1,\text{cascade}}$, the cascaded GAN loss $\mathcal{L}_{G, \text{cascade}}$ and face-id~\cite{deng2019arcface} loss $\mathcal{L}_\text{faceid}$ are calculated between the generated frame $\hat{I}_d$ and target frame $I_d$. The detailed definitions of these loss terms are available in the supplementary material. Moreover, due to the better disentanglement of facial expression and head pose, we do not need to train the second stage in LivePortrait~\cite{guo2024liveportrait}, including stitching and retargeting modules.

\begin{figure}[t!]
\vspace{-2mm}
  \centering
  \includegraphics[width=\columnwidth]{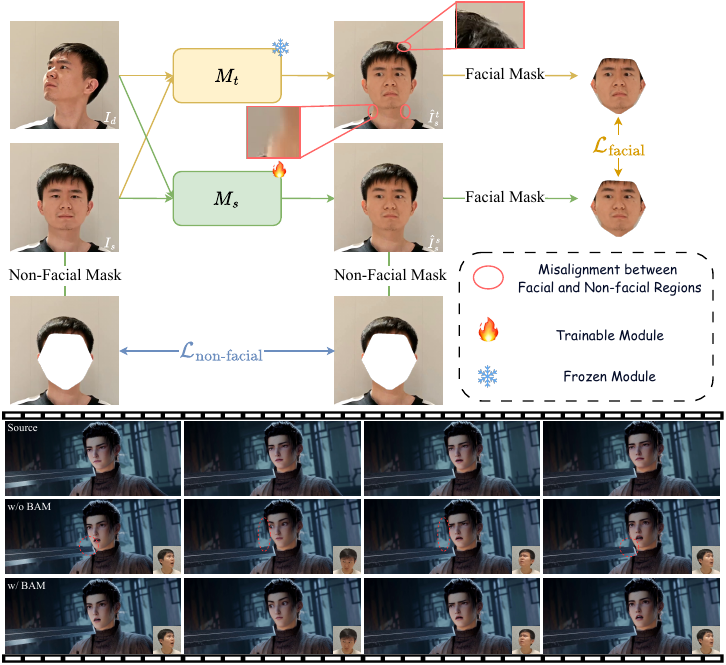}
  \caption{Top: we adopt the boundary alignment module to alleviate the misalignment between the facial and non-facial regions. Bottom: we provide some visual comparisons to demonstrate the necessity of using boundary alignment module (denoted as ``BAM"). The red dashed circles highlight the misalignment between the facial and non-facial regions. The bottom-right insets show the driving frames. Please zoom in for better inspection.}
  \vspace{-7mm}
  \label{fig:teacher}
\end{figure}

\subsection{Boundary Alignment Module}
Another important objective of portrait video expression editing is to maintain the smooth alignment around the boundary of face. However, to minimize the animation loss in ~\cref{equ:animate} on global region, the learned warping field of explicit 3D keypoints will also influence the non-facial region, resulting in misalignment on the boundary when performing facial expression editing task. To solve this problem, we design a boundary alignment module to mitigate the warping impact of explicit 3D keypoints on non-facial region, which is shown in the top of~\cref{fig:teacher}. Specifically, we firstly train a teacher model $M_t$ using the animation loss in ~\cref{equ:animate} between the generated image $\hat{I}_d^t$ and driving image $I_d$. Despite the misalignment between the facial and non-facial regions in generative results of $M_t$, it is capable of synthesizing precise and clear facial expression on the facial region. Therefore, given the source frame $I_s$ and target frame $I_d$, we first employ the trained teacher model $M_t$ to edit only the facial expression of $I_s$, obtaining an intermediate result $\hat{I}_s^t$. The keypoints of this process is calculated as:
\vspace{-2mm}
\setlength{\belowdisplayskip}{5pt}
\begin{equation}
    \left\{
        \begin{array}{lr}
        x_{s} = s_s \cdot \big((x_{c,s} +\delta_{s}) R_s \big) + t_{s}, &  \\
        x_{d} = s_s \cdot \big((x_{c,s} + \delta_{d}) R_s \big) + t_{s}, &
        \end{array}
    \right.
    \label{equ:replace}
\vspace{-1mm}
\end{equation}
which only replaces the expression deformation $\delta$ with that of $I_d$ while keeping all other parameters unchanged.

The student model $M_s$ is then trained to synthesize both the facial expression replacement results $\hat{I}_s^s$ and driving frame reconstruction result $\hat{I}_d^s$ from $I_s$. Additional loss terms are applied separately on each region of $\hat{I}_s^s$: for facial region, we compute a perceptual loss $\mathcal{L}_{P, \text{facial}}$ and an $L_1$ loss $\mathcal{L}_{1,\text{facial}}$ between $\hat{I}_s^s$ and $\hat{I}_s^t$; for non-facial region, we also calculate a perceptual loss $\mathcal{L}_{P, \text{non-facial}}$ and an $L_1$ loss $\mathcal{L}_{1,\text{non-facial}}$ between $\hat{I}_s^s$ and the ground truth $I_s$. These two losses can be formulated as follows:
\vspace{-3mm}
\begin{align}
& \mathcal{L}_\text{facial} = \mathcal{L}_{P, \text{facial}} + \mathcal{L}_{1,\text{facial}}, \\ 
& \mathcal{L}_\text{non-facial} = \mathcal{L}_{P, \text{non-facial}} + \mathcal{L}_{1,\text{non-facial}},
\end{align}
Meanwhile, the animation loss in~\cref{equ:animate} is also applied between the generated frame $\hat{I}_d^s$ and driving frame $I_d$. We provide the detailed calculation process for the mask of facial and non-facial regions in the supplementary material. The bottom part of~\cref{fig:teacher} shows some visual comparisons of our model trained with or without boundary alignment module, highlighting its importance in training pipeline.

\subsection{Inference}
Given a source video sequence $\{I_{s,i}|i=0,1,...,N-1\}$ and a driving video sequence $\{I_{d,i}|i=0,1,...,N-1\}$, we propose two modes: replacement mode and enhancement mode for portrait video expression editing in the inference stage. Replacement mode directly replace the facial expression of the $i$-th frame $I_{s,i}$ in source video with that of the $i$-th frame $I_{d,i}$ in driving video. The keypoints transformation of this mode is the same as~\cref{equ:replace}.

Enhancement mode aims to enhance the facial expression in source video, whose keypoint transformation process of the $i$-th frame is modified to:
\begin{equation}
\resizebox{0.9\linewidth}{!}{$
    \left\{
        \begin{array}{lr}
        x_{s,i} = s_{s,i} \cdot \big((x_{c,s} +\delta_{s,i}) R_{s,i} \big) + t_{s,i}, &  \\
        x_{d,i} = s_{s,i} \cdot \big((x_{c,s} + \delta_{s,i} + \delta_{d,i} - \delta_{d,0}) R_{s,i} \big) + t_{s,i}. &
        \end{array}
    \right.
    $}
    \label{equ:enhance}
\end{equation}
which adds the facial expression of driving video on the top of that of source video. The appearance feature volume $f_{s,i} = \mathcal{F}(I_{s,i})$ is extracted from the $i$-th frame of source video in both modes. For portrait animation task, the keypoints transformation in inference stage is the same as~\cref{equ:keypoint} in training stage. We provide more details in the supplementary material.
\vspace{-1mm}
\section{Experiments}
\label{sec:expr}
\begin{figure*}[t!]
  \centering
  \includegraphics[width=0.95\linewidth]{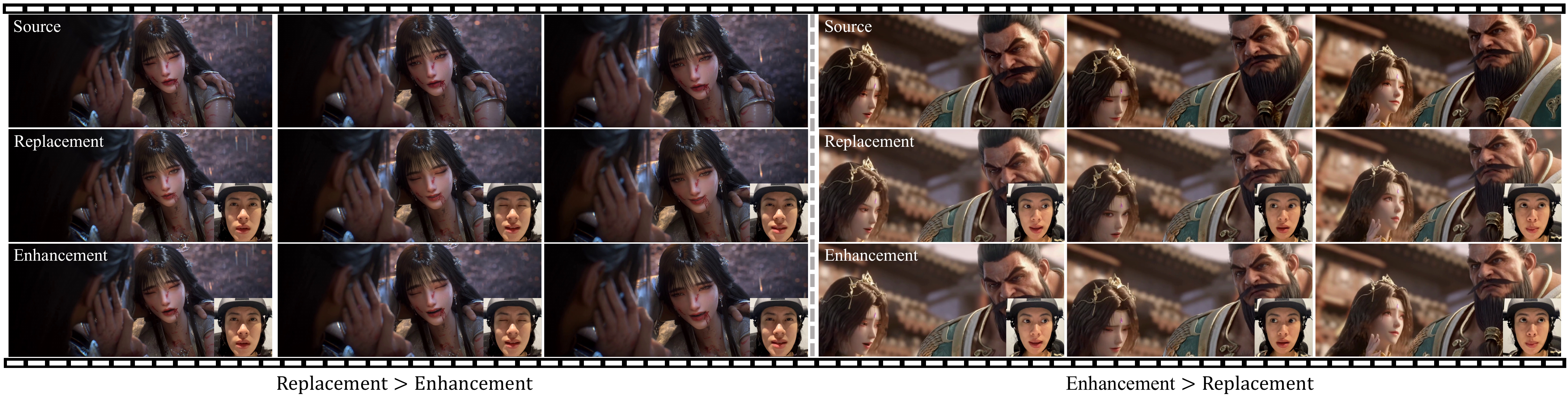}
  \vspace{-2mm}
  \caption{The typical usage scenarios for our proposed replacement mode and enhancement mode. The bottom-right insets exhibit driving frames. Please zoom in for better inspection.}
  \label{fig:two_infer_modes}
  \vspace{-6mm}
\end{figure*}

\vspace{-1mm}
\paragraph{Implementation Details.}
We adopt a similar architecture to LivePortrait~\cite{guo2024liveportrait}, except that our appearance feature extractor $\mathcal{F}$ is built upon the pretrained DINOv2~\cite{oquab2023dinov2} backbone. To calculate the FLAME loss $\mathcal{L}_\text{FLAME}$, the number of explicit 3D keypoints $K$ is set to 49, which cover most of the key regions of the face. The resolution of input frames and output image of our model is set to $512 \times 512$. The details of keypoints selection and training settings are presented in the supplementary material.

\vspace{-5mm}
\paragraph{Dataset.}
We utilized a combination of several publicly-available datasets, including VFHQ\cite{xie2022vfhq}, MEAD\cite{kaisiyuan2020mead}, Nersemble\cite{kirschstein2023nersemble}, FEED\cite{drobyshev2024emoportraits}, and ETH-XGaze\cite{Zhang2020ETHXGaze}. 
We also incorporate a large amount of portrait videos from the Internet, including high-definition anime and film, to further enhance the diversity and quality of our in-house dataset. Finally, this results in a total of 597,331 video clips, which include a variety of expressions and emotional intensities.

\vspace{-5mm}
\paragraph{Evaluation Metrics.}
To measure the generation quality and controllability of our PerformRecast on both video expression editing and portrait animation task, we adopt PSNR, SSIM~\cite{wang2004image}, LPIPS~\cite{zhang2018unreasonable}, $\mathcal{L}_1$ distance, Fréchet Inception Distance (FID)~\cite{heusel2017gans},  Fréchet Video Distance (FVD)~\cite{unterthiner2019fvd}, Cosine SIMilarity of identity features (CSIM)~\cite{kim2022adaface}, Average Expression Distance (AED)~\cite{siarohin2019first}, Average Pose Distance (APD)~\cite{siarohin2019first} and Mean Angular Error (MAE) of eyeball direction~\cite{han2024face}. Details of these metrics are provided in the supplementary material. All these metrics are calculated under the resolution of $512 \times 512$ for each compared method.

\begin{figure}[t!]
  \centering
  \includegraphics[width=\columnwidth]{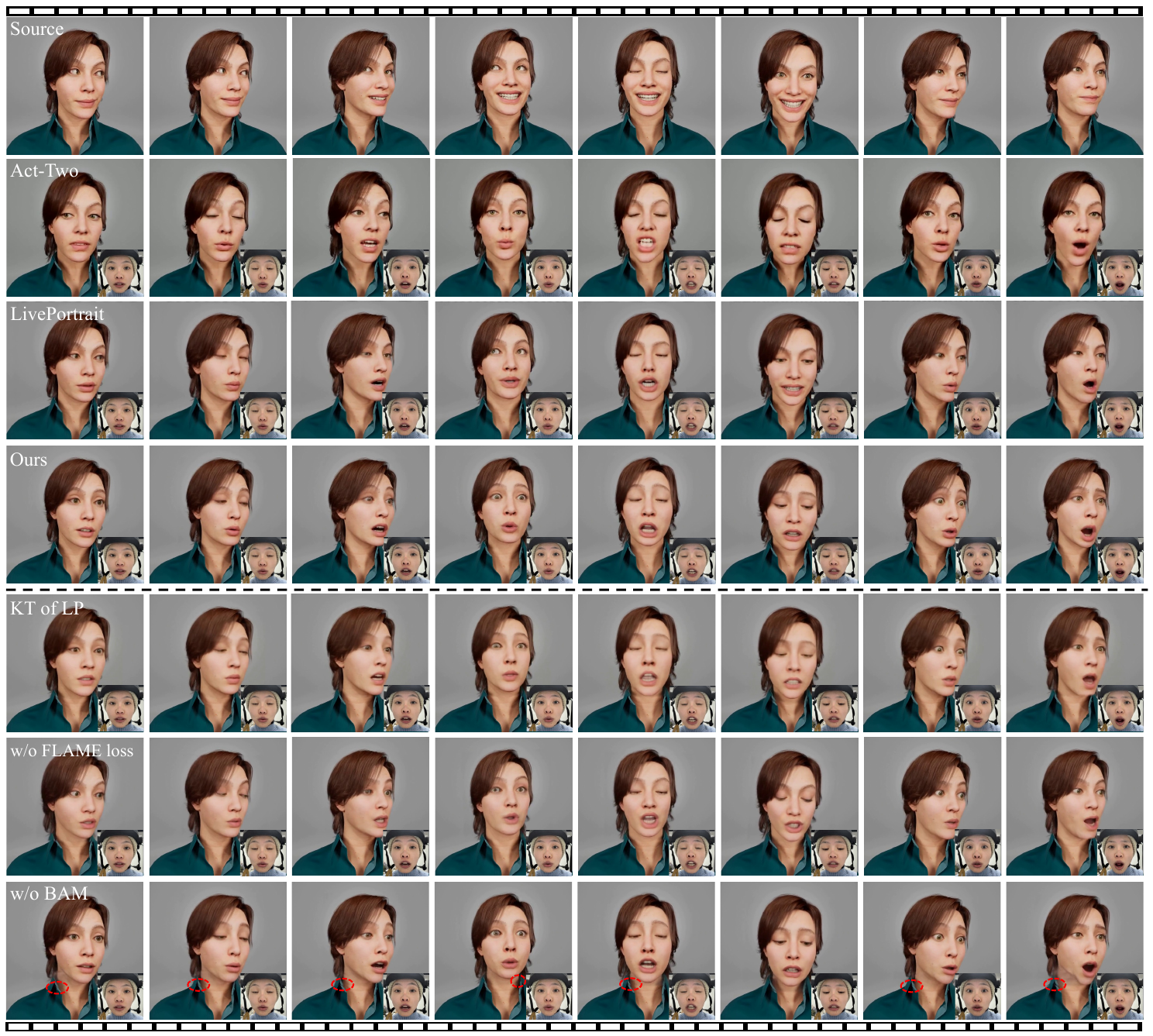}
  \vspace{-6mm}
  \caption{Qualitative comparison of portrait video expression editing on replacement mode. The top of the figure shows editing results of different methods. The bottom presents our ablation studies and analysis. The bottom-right insets exhibit driving frames. The red dashed circles highlight the misalignment between the facial and non-facial regions. Please zoom in for better inspection.}
  \vspace{-6mm}
  \label{fig:replacement}
\end{figure}

\begin{table*}[t]
  \centering
  \caption{Quantitative results of facial expression editing on our test benchmark on both replacement and enhancement modes. The top of the table shows the comparison with other methods while the bottom presents our ablation studies and analysis.}
  \vspace{-2mm}
  \resizebox{\linewidth}{!}{
  \begin{tabular}{lcccccccccccccccccccc}
      \toprule[1.5pt]
      \multirow{2}[2]{*}{Method}  & \multicolumn{10}{c}{\textbf{Replacement Mode}} & \multicolumn{10}{c}{\textbf{Enhancement Mode}} \\ 
      \cmidrule[0.5pt](lr){2-11} \cmidrule[0.5pt](lr){12-21}
       & PSNR$\uparrow$ & SSIM$\uparrow$ & LPIPS$\downarrow$ & $\mathcal{L}_1$$\downarrow$ & CSIM$\uparrow$ & MAE(\textdegree)$\downarrow$ &AED$\downarrow$ &APD$\downarrow$ &FID$\downarrow$ &FVD$\downarrow$ & PSNR$\uparrow$ & SSIM$\uparrow$ & LPIPS$\downarrow$ & $\mathcal{L}_1$$\downarrow$ & CSIM$\uparrow$ & MAE(\textdegree)$\downarrow$ &AED$\downarrow$ &APD$\downarrow$ &FID$\downarrow$ &FVD $\downarrow$\\
      \midrule[1pt]
       
       SkyReels-A1~\cite{qiu2025skyreels}  & 24.9051 & 0.8586 & 0.1622 & 0.0418 & 0.7163 & 13.0791 & 0.7157 & 0.0155 & 50.2007 & 1249.7021 & 24.9206 & 0.8584 & 0.1612 & 0.0417 & 0.7176 & 11.3797 & 0.676 & 0.016 & 51.47 & 1196.1767\\
       
       Hunyuan-Portrait~\cite{xu2025hunyuanportrait}  &22.4287 & 0.7924 & 0.1691 & 0.0423 & 0.7364 & 10.4001 & 0.6608 & 0.0348 & 38.5997 & 1925.0654 & 22.7968 & 0.7984 & 0.1649 & 0.0406 & 0.736 & 9.6267 & 0.6073 & 0.0332 & 39.3112 & 1861.9558\\
       
       FantasyPortrait~\cite{wang2025fantasyportrait}  & 23.9456 & 0.8204 & 0.1883 & 0.0349 & 0.7338 & 16.7765 & 0.7953 & 0.0168 & 60.4381 & 606.6303 & 24.2032 & 0.8239 & 0.1848 & 0.0339 & 0.7387 & 12.2154 & 0.6614 & 0.0155 & 62.0027 & 551.448 \\
       
       Wan-Animate~\cite{cheng2025wan}  &22.8196 & 0.8021 & 0.1319 & 0.0467 & 0.614 & 11.9722 & 0.7002 & 0.0238 & 28.1003 & 849.2188  & 22.6417 & 0.8006 & 0.1402 & 0.0491 & 0.6062 & 11.2752 & 0.6847 & 0.0225 & 31.2186 & 885.4313\\

      Act-Two~\cite{runway}  & 20.8344 & 0.7913 & 0.1634 & 0.0459 & 0.6819 & 15.9618 & 0.7901 & 0.0723 & 36.1933 & 322.0726  & 20.8066 & 0.7916 & 0.165 & 0.0458 & 0.685 & 15.4109 & 0.8019 & 0.0727 & 38.0579 & 330.5131\\
      
     LivePortrait~\cite{guo2024liveportrait}  &27.7296 & \underline{0.8989} & 0.0591 & 0.0183 & 0.7494 & 10.4746 & 0.6098 & 0.0162 & \underline{14.3562} & 165.1011 &  28.0103 & 0.9024 & 0.0479 & 0.0172 & 0.796 & 7.6325 & 0.4915 & 0.0112 & 12.2974 & 114.2545\\

     \textbf{Ours}  & \textbf{29.2724} & \textbf{0.9141} & \textbf{0.0474} & \textbf{0.014} & \textbf{0.7613} & 9.1217 & \textbf{0.4986} & \textbf{0.0122} & \textbf{12.0138} & \textbf{102.9898} &  \textbf{30.2665} & \textbf{0.9216} & \textbf{0.0394} & \textbf{0.0128} & \underline{0.8191} & \textbf{6.8211} & \underline{0.4472} & \textbf{0.0102} & \underline{10.7694} & \textbf{90.2483}  \\
     
    \midrule
    
    Ours (KT of LP) & 27.0623 & 0.8908 & 0.0968 & 0.018 & \underline{0.7512} & \textbf{8.7438} & \underline{0.5733} & \underline{0.0144} & 27.6785 & 288.8396 &  28.3379 & 0.903 & 0.0548 & 0.0151 & 0.785 & 7.2571 & 0.473 & 0.013 & 13.8178 & 139.0848\\
    
    Ours (w/o FLAME loss) &  24.9869 & 0.8726 & 0.0805 & 0.0224 & 0.7268 & 9.9748 & 0.6625 & 0.0324 & 18.3957 & 188.1082  & 28.4225 & 0.8977 & 0.0838 & 0.0159 & 0.8063 & 9.728 & 0.5384 & 0.0132 & 19.5149 & 132.4458 \\

    Ours (w/o T-S) & \underline{27.7346} & 0.8976 & \underline{0.0583} & \underline{0.0166} & 0.7395 & \underline{8.847} & 0.5749 & 0.0147 & 14.4061 & \underline{136.4097} &  \underline{29.6156} & \underline{0.9148} & \underline{0.0421} & \underline{0.0135} & \textbf{0.821} & \underline{7.025} & \textbf{0.4453} & \underline{0.0106} & \textbf{10.7291} & \underline{100.7387}\\
    
      \bottomrule[1.5pt]
  \end{tabular}
  }
  \vspace{-6mm}
  \label{tab:expression}
\end{table*}

\vspace{-1mm}
\subsection{Portrait Video Expression Editing}

We first would like to clarify the two inference modes and discuss their respective usage scenarios. When the facial expression in a shot does not fully satisfy the director’s creative expectations, two inference modes arise. Replacement mode is appropriate when the actor’s overall performance is compelling and remains consistent with the narrative context. Enhancement mode is more suitable when only localized improvements are needed without compromising the performance plausibility. Fig.~\ref{fig:two_infer_modes} shows the typical usage scenarios for our proposed two inference modes.

We then conduct experiments to compare the ability of our proposed PerformRecast with other methods on portrait video expression editing task. Besides LivePortrait~\cite{guo2024liveportrait}, we modify several diffusion-based portrait animation methods, including SkyReels-A1~\cite{qiu2025skyreels}, Hunyuan-Portrait~\cite{xu2025hunyuanportrait}, FantasyPortrait~\cite{wang2025fantasyportrait} and Wan-Animate~\cite{cheng2025wan} to make them capable of editing only the expressions of source video according to driving video. Specifically, we combine the expression information of driving frame with the head pose information of source frame to generate the editing results. The detailed modification of each method is described in the supplementary material. We also compare with the closed-source commercial product Runway Act-Two~\cite{runway}.

To more accurately evaluate the performance of different methods, we construct a test benchmark utilizing MetaHuman~\cite{metahuman}. Our test benchmark contains 18 portrait videos with diverse expression and without head pose rotation recorded from professional facial motion actors. Then, we select 20 digital humans from MetaHuman, and for each digital human, we render 19 videos, of which 18 are using the expressions of facial motion actors, and the last one is without expression. In addition, all the videos are added with our pre-defined head pose rotation. Further details of our test benchmark can be found in the supplementary material. 
For replacement mode, we randomly select one of the 18 expressions as source video, and select another different expression from facial motion actors as driving video for each digital human. For enhancement mode, we select the video without expression as source video. As a result, for each mode, we obtain 20 source-driving-ground truth video triplets, one for each digital human. We will release our constructed test benchmark to facilitate the future research.

\vspace{-5mm}
\paragraph{Qualitative results.}
The top of ~\cref{fig:replacement} presents some qualitative results generated by different methods on replacement modes. LivePortrait~\cite{guo2024liveportrait} performs poorly and tends to generate inaccurate eyeballs direction. Act-Two~\cite{runway} fails to accurately preserve the head pose of source video and struggles to synthesize the fine-grained expression details from the driving video. On the contrary, our method is capable of faithfully preserving both the head pose of source video and facial expressions in driving video. More visual comparisons on the enhancement mode are provided in the supplementary material.

\vspace{-5mm}
\paragraph{Quantitative results.}
We provide the quantitative comparison with different methods at the top part of~\cref{tab:expression}. Due to the delicate design of expression and head pose disentanglement, our method outperforms all other methods on all metrics. Diffusion-based methods inherently lack the capability of expression-only video editing, and their performances are still of low quality even after our modification. Therefore, we do not conduct qualitative comparison with them in the main manuscript.

\vspace{-1mm}
\subsection{Ablation Studies and Analysis}

\begin{figure}[t!]
  \centering
  \includegraphics[width=\columnwidth]{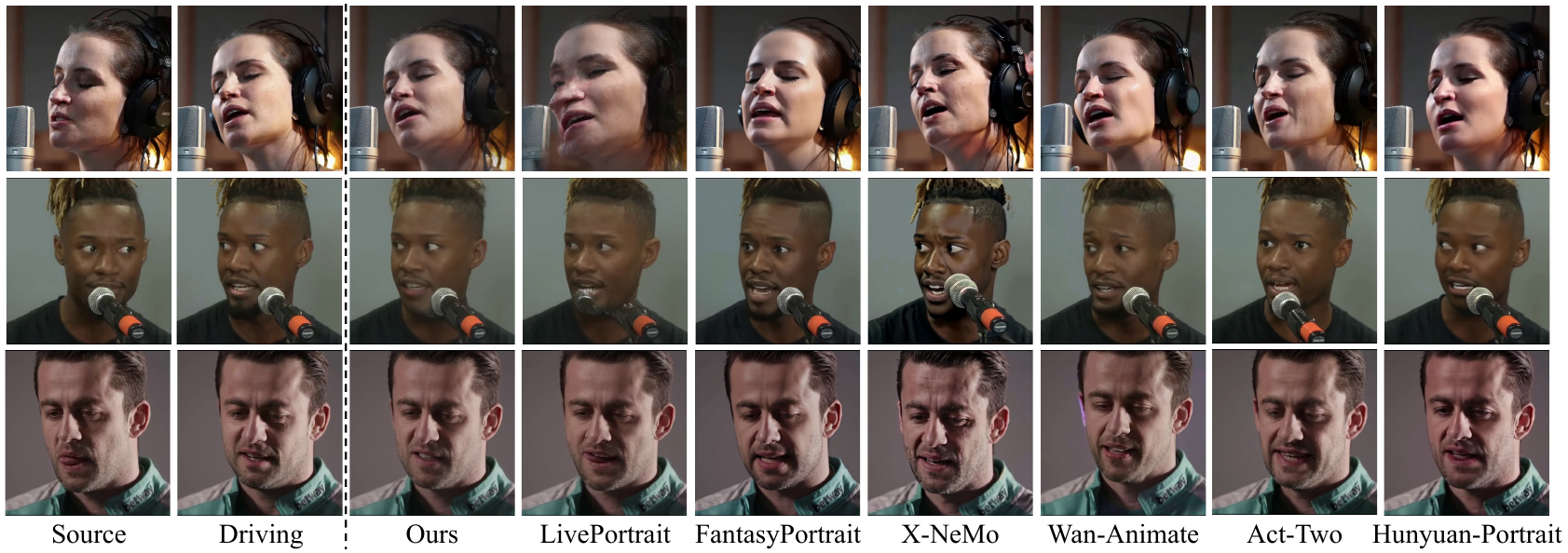}
  \vspace{-6mm}
  \caption{Qualitative comparison of self-reenactment portrait animation. We show the source frame, driving frame and generative results. These source-driving paired images are from 
  official test split of VFHQ dataset~\cite{xie2022vfhq}. Please zoom in for better inspection.}
  \vspace{-6mm}
  \label{fig:self-id}
\end{figure}

\begin{table*}[t]
  \centering
  \caption{Quantitative comparisons of different methods on portrait animation task. The top of the table shows the results of non-diffusion-based methods while the bottom presents diffusion-based methods.
  }
  \vspace{-3mm}
  \resizebox{0.93\linewidth}{!}{
  \begin{tabular}{lcccccccccccccc}
      \toprule[1.5pt]
      \multirow{2}[2]{*}{Method}  & \multicolumn{10}{c}{\textbf{Self-reenactment}} & \multicolumn{4}{c}{\textbf{Cross-reenactment}} \\ 
      \cmidrule[0.5pt](lr){2-11} \cmidrule[0.5pt](lr){12-15}
       & PSNR$\uparrow$ & SSIM$\uparrow$ & LPIPS$\downarrow$ & $\mathcal{L}_1$$\downarrow$ & CSIM$\uparrow$ & MAE(\textdegree)$\downarrow$ &AED$\downarrow$ &APD$\downarrow$ &FID$\downarrow$ &FVD$\downarrow$ & 
       
       CSIM$\uparrow$ & MAE(\textdegree)$\downarrow$ &AED$\downarrow$ &APD$\downarrow$\\
      \midrule[1pt]

       GAGAvatar~\cite{chu2024gagavatar} & - & - & - & - & 0.7878& 8.1317 & 0.4057 & 0.014 & - & - & 0.6871 & 12.2879 & 0.8144 & \underline{0.0256} \\
       
       Portrait4D-v2~\cite{deng2024portrait4dv2} &14.3781 & 0.5601 & 0.4877 & 0.1268 & 0.7525 & 8.4331 & 0.4434 & 0.0262 & 55.0744 & 506.1298 &0.6702& 13.4384 & 0.8385 & 0.0321 \\
       
       PD-FGC~\cite{wang2022pdfgc}  & 16.4477 & 0.62 & 0.4252 & 0.0936 &0.3282& 11.9368 & 0.6345 & 0.0322 & 76.2318 & 628.5279 &0.2181 & 15.0314 & 0.8988 & 0.0417 \\
       
       EMOPortrait~\cite{drobyshev2024emoportraits} &18.7677 & 0.6979 & 0.2711 & 0.0648 & 0.6297 & 7.7967 & 0.4676 & 0.0179 & 43.2991 & 444.5448 & 0.3483 & \textbf{10.3905} & 0.8012 & \textbf{0.0245} \\
       
       EDTalk~\cite{tan2024edtalk} & 20.0771 & 0.7272 & 0.3054 & 0.0648 & 0.674 & 16.3315 & 0.572 & 0.0252 & 60.5285 &  369.9713 & 0.5435 & 22.5909 & 1.018 & 0.0531\\

       LIA-X~\cite{wang2025lia} & 18.249 & 0.6711 & 0.2763 & 0.0731 & 0.7416 & 10.6918 & 0.5797 & 0.0768 & 35.1192 &  317.5413 & \textbf{0.8270} & 27.6757 & 1.2581 & 0.2083\\
       
       LivePortrait~\cite{guo2024liveportrait} & \textbf{22.8809} & \underline{0.7891} & \underline{0.165} & \underline{0.0433} & \underline{0.8008} & \underline{6.595} & \underline{0.3419} & \underline{0.0095} & \underline{21.3192} & \underline{192.0196} & 0.6595 & 12.6259 & 0.8264 & 0.0295 \\
       
       \midrule[1pt]

       FYE~\cite{ma2024follow}  & 20.1905 & 0.7168 & 0.2118 & 0.0564 & 0.7618 & 11.8316 & 0.5676 & 0.0273 & 30.0686 & 343.32 & 0.7187 & 15.4403 & 1.1178 & 0.0482 \\
       
       AniPortrait~\cite{wei2024aniportrait} & 21.0342 & 0.7334 & 0.1809 & 0.0499 & 0.7654 & 10.0084 & 0.4143 & 0.015 & 26.6765 & 210.9606 & 0.6894 & 18.6912 & 1.117 & 0.0444 \\
       
       X-NeMo~\cite{zhao2025x} & 16.5234 & 0.5666 & 0.3404 & 0.096 & 0.7472 & 10.3959 & 0.4102 & 0.0152 & 34.3514 & 373.0014 & 0.6555 & 12.8173 & \underline{0.7997} & 0.028 \\
       
       ReliPA~\cite{guo2025high} &16.1173 & 0.6264 & 0.3702 & 0.103 & 0.5317 & 12.5351 & 0.5572 & 0.021 & 41.1397 & 470.7546 & 0.553 & 28.4037 & 1.2146 & 0.199 \\
       
       SkyReels-A1~\cite{qiu2025skyreels} & 17.4286 & 0.6644 & 0.3332 & 0.087 & 0.7103 & 13.1925 & 0.5291 & 0.0306 & 34.9209 & 363.1019 & 0.5856 & 21.6549 & 1.0562 & 0.1058 \\
       
       Hunyuan-Portrait~\cite{xu2025hunyuanportrait} & 16.97 & 0.6366 & 0.3291 & 0.0873 & 0.7741 & 9.9504 & 0.4218 & 0.0355 & 28.0526 & 266.6914 & 0.5939 & 17.7903 & 0.9142 & 0.0899 \\
       
       FantasyPortrait~\cite{wang2025fantasyportrait} & 16.8794 & 0.6321 & 0.3394 & 0.0963 &0.7368 & 9.9752 & 0.4743 & 0.0393 & 42.5995 & 446.281 & \underline{0.7694} & 18.581 & 0.9745 & 0.1475 \\
       
       Wan-Animate~\cite{cheng2025wan} &18.3191 & 0.6505 & 0.2825 & 0.0768 & 0.7327 & 9.5602 & 0.4797 & 0.0186 & 26.8861 & 302.6864 & 0.5812 & 14.7004 & 0.9231 & 0.0405 \\
       
       VACE~\cite{vace} & 11.1789 & 0.5398 & 0.5259 & 0.2117 & 0.4311 & 13.0378 & 0.6314 & 0.0248 & 99.9335 & 918.8369 & 0.4001 & 19.5828 & 0.9737 & 0.0396 \\
       
       AvatarArtist~\cite{liu2025avatarartist} &13.3046 & 0.5414 & 0.5791 & 0.1524 &0.4025 & 16.3325 & 0.6983 & 0.0443 & 91.3534 & 1043.3879 & 0.5073 & 14.8549 & 0.9374 & 0.0338 \\ 
       
       \midrule[1pt]

      \textbf{Ours} &  \underline{22.7117} & \textbf{0.7895} & \textbf{0.1593} & \textbf{0.0409} & \textbf{0.8434} & \textbf{4.9976} & \textbf{0.2606} & \textbf{0.009} & \textbf{20.1612} & \textbf{164.1895} & 0.6966 & \underline{10.9564} & \textbf{0.7025} & 0.0303 \\

      \bottomrule[1.5pt]
  \end{tabular}
  }
  \vspace{-3mm}
  \label{tab:animation}
\end{table*}

\vspace{-1mm}
\paragraph{Keypoints Transformation Formula.}
We first analyze the importance of using our improved keypoints transformation instead of that of LivePortrait. We train an additional model using the keypoints transformation of LivePortrait on expression editing task. The third-to-last row of~\cref{fig:replacement} and~\cref{tab:expression} (denoted as ``KT of LP'') shows the generated results. This variant tends to synthesize relatively blurry videos, and exhibits suboptimal quantitative metrics.

\vspace{-5mm}
\paragraph{Essentials of FLAME Loss.} We then verify the effectiveness of FLAME loss $\mathcal{L}_\text{FLAME}$ in our training pipeline by training a PerformRecast variant without FLAME loss, which is reported at the second-to-last row of~\cref{fig:replacement} and \cref{tab:expression}. Without the supervision from FLAME loss, the performance of our model drops markedly, due to the inaccuracy of the motion extractor.

\begin{figure*}[t!]
  \centering
  \includegraphics[width=0.95\linewidth]{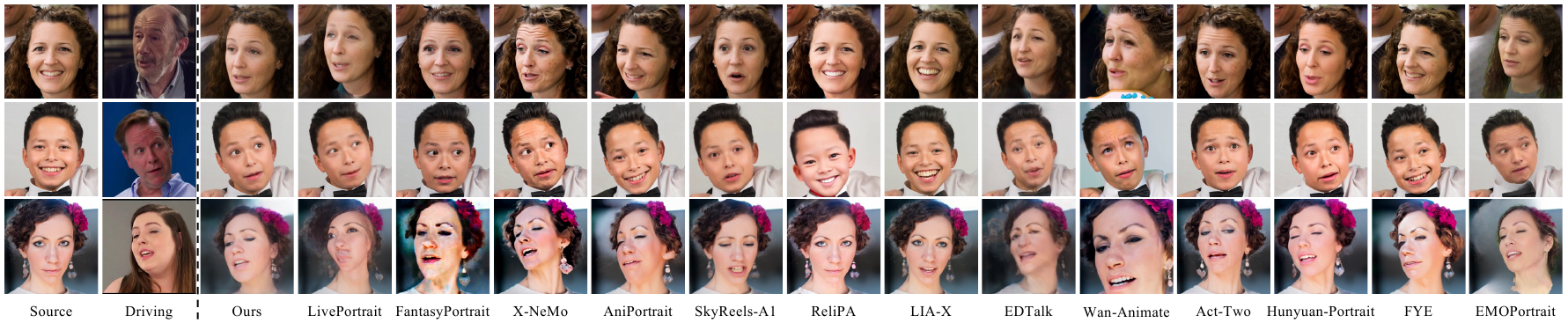}
  \vspace{-3mm}
  \caption{Qualitative comparison of cross-reenactment portrait animation. We show the source image, driving image and the generative results. The source images are from FFHQ dataset and driving frames are from VFHQ dataset.}
  \label{fig:cross-id}
\vspace{-6mm}
\end{figure*}

\vspace{-4mm}
\paragraph{Essentials of Boundary Alignment Module.} Finally, we provide the performance of our model trained without boundary alignment module at the last row of ~\cref{tab:expression} and~\cref{fig:replacement} (denoted as ``w/o BAM''). From which we can conclude that boundary alignment module effectively alleviates the misalignment between the facial and non-facial regions in generated images and leads to the improvement on pixel-level evaluation metrics (such as PSNR, FID, etc.).

\vspace{-2mm}
\subsection{Portrait Animation}
We further investigate our model's ability on portrait animation task, which aims to animate a static portrait using both the facial expression and head pose from a driving video. We compare our model with several non-diffusion-based methods, including GAGAvatar~\cite{chu2024gagavatar}, EDTalk~\cite{tan2024edtalk}, LivePortrait~\cite{guo2024liveportrait}, together with other approaches mentioned in~\cref{sec:non-diffusion}. We also compare against recent diffusion-based models, such as AniPortrait~\cite{wei2024aniportrait}, X-NeMo~\cite{zhao2025x}, Wan-Animate~\cite{cheng2025wan}, as well as some other methods in~\cref{sec:diffusion}. Finally, we compare our model with Act-Two~\cite{runway}.

\vspace{-1mm}
\subsubsection{Self-reenactment}
Self-reenactment portrait animation task uses the first frame as the source image and animate it using the whole frames in the same video. All the methods are evaluated on official test split of VFHQ dataset~\cite{xie2022vfhq}, which consists of 50 videos.

\vspace{-4mm}
\paragraph{Qualitative results.}
~\cref{fig:self-id} provides some qualitative results on the same source-driving paired frames by different methods. LivePortrait~\cite{guo2024liveportrait} is likely to move other object in the source frame, such as the microphone in the second case. Other diffusion-based methods tend to produce unstable results and struggle to capture subtle facial expression, such as eye gazes and lip movements. On the contrary, our model faithfully reconstructs the fine-grained expressions and head pose of the driving frame.

\vspace{-5mm}
\paragraph{Quantitative results.}
The left part of~\cref{tab:animation} reports the comparison of each metric between PerformRecast and other methods on self-reenactment. Since GAGAvatar~\cite{chu2024gagavatar} can only produce results with black background, we do not calculate pixel-level metrics for it. From~\cref{tab:animation} we can observe that our method obtains the best performance on all metrics except PSNR, outperforming all other methods.

\vspace{-1mm}
\subsubsection{Cross-reenactment}
\vspace{-1mm}
For cross-reenactment portrait animation, we select 50 images from FFHQ dataset~\cite{karras2019style} as source portraits. The official test split of VFHQ dataset~\cite{xie2022vfhq} are used as driving videos. We only adopt CSIM, MAE, AED and APD as evaluation metrics due to the lack of ground truth target images.

\vspace{-5mm}
\paragraph{Qualitative results.}
~\cref{fig:cross-id} visualizes the generated results of different compared methods on the same source-driving paired images. LivePortrait~\cite{guo2024liveportrait} exhibits severe artifacts on the third case. Other diffusion-based methods generally produce overly exaggerated facial expression, such as wrinkles on the forehead and the wide smiles around the lips. In contrast, our method is capable of faithfully transferring the subtle facial expression, direction of eyeballs and head pose rotations from the driving image to source image. 

\vspace{-5mm}
\paragraph{Quantitative results.}
The right part of ~\cref{tab:animation} presents the quantitative results of the cross-reenactment comparisons. Our model outperforms all previous methods in terms of facial expression accuracy. While EMOPortrait~\cite{drobyshev2024emoportraits} reports lower MAE and APD than ours, its generated results suffer from severe background blur and distortion as shown in~\cref{fig:cross-id}. What's more, although FantasyPortrait~\cite{wang2025fantasyportrait} and FYE~\cite{ma2024follow} attain higher facial identity similarity, they lag significantly behind on all other metrics. Therefore, we can conclude that our PerformRecast achieves state-of-the-art overall performance on the cross-reenactment task. In addition, our method is able to generate six images per second on the consumer-grade GPU device, which significantly reduces the complexity of deployment in practical scenarios.
\vspace{-7mm}
\section{Conclusion}
\label{sec:conclusion}
\vspace{-2mm}
In this paper, we propose PerformRecast, a versatile and effective method tailored for expression-only video performance editing and portrait animation tasks. We improve the original keypoints transformation formula in LivePortrait~\cite{guo2024liveportrait} to make it more consistent with 3DMM. As a result, our PerformRecast is much better at disentangling the face identity, facial expression and head pose rotation. Experimental results demonstrate that our method is capable of editing the expression of a portrait video as well as animating a static portrait image, offering great convenience in the film and computer animation industries.

\small
\bibliographystyle{ieeenat_fullname}
\bibliography{main}

@String(CVPR= {IEEE Conf. Comput. Vis. Pattern Recog.})

@String(ICCV= {Int. Conf. Comput. Vis.})

@String(ECCV= {Eur. Conf. Comput. Vis.})

@String(TOG= {ACM Trans. Graph.})

@String(AAAI = {AAAI})

@String(CVPRW= {IEEE Conf. Comput. Vis. Pattern Recog. Worksh.})

@String(CVPR  = {CVPR})

@String(ICCV  = {ICCV})

@String(ECCV  = {ECCV})

@String(TOG   = {ACM TOG})

@String(CVPRW= {CVPRW})

@misc{giebenhain2025pixel3dmm,
title={Pixel3DMM: Versatile Screen-Space Priors for Single-Image 3D Face Reconstruction},
author={Simon Giebenhain and Tobias Kirschstein and  Martin R{\"{u}}nz and Lourdes Agapito and Matthias Nie{\ss}ner},
year={2025},
url={https://arxiv.org/abs/2505.00615},
}

@article{goodfellow2014generative,
  title={Generative adversarial nets},
  author={Goodfellow, Ian J and Pouget-Abadie, Jean and Mirza, Mehdi and Xu, Bing and Warde-Farley, David and Ozair, Sherjil and Courville, Aaron and Bengio, Yoshua},
  journal={Advances in neural information processing systems},
  volume={27},
  year={2014}
}

@inproceedings{karras2020analyzing,
  title={Analyzing and improving the image quality of stylegan},
  author={Karras, Tero and Laine, Samuli and Aittala, Miika and Hellsten, Janne and Lehtinen, Jaakko and Aila, Timo},
  booktitle={Proceedings of the IEEE/CVF conference on computer vision and pattern recognition},
  pages={8110--8119},
  year={2020}
}

@inproceedings{rombach2022high,
  title={High-resolution image synthesis with latent diffusion models},
  author={Rombach, Robin and Blattmann, Andreas and Lorenz, Dominik and Esser, Patrick and Ommer, Bj{\"o}rn},
  booktitle={Proceedings of the IEEE/CVF conference on computer vision and pattern recognition},
  pages={10684--10695},
  year={2022}
}

@article{ho2020denoising,
  title={Denoising diffusion probabilistic models},
  author={Ho, Jonathan and Jain, Ajay and Abbeel, Pieter},
  journal={Advances in neural information processing systems},
  volume={33},
  pages={6840--6851},
  year={2020}
}

@article{song2020denoising,
  title={Denoising Diffusion Implicit Models},
  author={Song, Jiaming and Meng, Chenlin and Ermon, Stefano},
  journal={arXiv:2010.02502},
  year={2020},
  month={October},
  abbr={Preprint},
  url={https://arxiv.org/abs/2010.02502}
}

@inproceedings{Drobyshev22MP,
    author = {Drobyshev, Nikita and Chelishev, Jenya and Khakhulin, Taras and Ivakhnenko, Aleksei and Lempitsky, Victor and Zakharov, Egor},
    title = {MegaPortraits: One-shot Megapixel Neural Head Avatars},
    journal   = {Proceedings of the 30th ACM International Conference on Multimedia},
    year      = {2022},
}

@misc{SD1-5,
  title        = {Stable diffusion v1.5 model card},
  author       = {Stability AI},
  year         = {2022},
  howpublished = {https://huggingface.co/stable-diffusion-v1-5/stable-diffusion-v1-5},
}

@article{blattmann2023stable,
  title={Stable video diffusion: Scaling latent video diffusion models to large datasets},
  author={Blattmann, Andreas and Dockhorn, Tim and Kulal, Sumith and Mendelevitch, Daniel and Kilian, Maciej and Lorenz, Dominik and Levi, Yam and English, Zion and Voleti, Vikram and Letts, Adam and others},
  journal={arXiv preprint arXiv:2311.15127},
  year={2023}
}

@article{yang2024cogvideox,
  title={CogVideoX: Text-to-Video Diffusion Models with An Expert Transformer},
  author={Yang, Zhuoyi and Teng, Jiayan and Zheng, Wendi and Ding, Ming and Huang, Shiyu and Xu, Jiazheng and Yang, Yuanming and Hong, Wenyi and Zhang, Xiaohan and Feng, Guanyu and others},
  journal={arXiv preprint arXiv:2408.06072},
  year={2024}
}

@article{hong2022cogvideo,
  title={CogVideo: Large-scale Pretraining for Text-to-Video Generation via Transformers},
  author={Hong, Wenyi and Ding, Ming and Zheng, Wendi and Liu, Xinghan and Tang, Jie},
  journal={arXiv preprint arXiv:2205.15868},
  year={2022}
}

@article{ma2024follow,
  title={Follow-Your-Emoji: Fine-Controllable and Expressive Freestyle Portrait Animation},
  author={Ma, Yue and Liu, Hongyu and Wang, Hongfa and Pan, Heng and He, Yingqing and Yuan, Junkun and Zeng, Ailing and Cai, Chengfei and Shum, Heung-Yeung and Liu, Wei and others},
  journal={arXiv preprint arXiv:2406.01900},
  year={2024}
}

@article{zhao2025x,
  title={X-NeMo: Expressive neural motion reenactment via disentangled latent attention},
  author={Zhao, Xiaochen and Xu, Hongyi and Song, Guoxian and Xie, You and Zhang, Chenxu and Li, Xiu and Luo, Linjie and Suo, Jinli and Liu, Yebin},
  journal={arXiv preprint arXiv:2507.23143},
  year={2025}
}

@article{cheng2025wan,
  title={Wan-Animate: Unified Character Animation and Replacement with Holistic Replication},
  author={Cheng, Gang and Gao, Xin and Hu, Li and Hu, Siqi and Huang, Mingyang and Ji, Chaonan and Li, Ju and Meng, Dechao and Qi, Jinwei and Qiao, Penchong and others},
  journal={arXiv preprint arXiv:2509.14055},
  year={2025}
}

@article{wang2025fantasyportrait,
  title={FantasyPortrait: Enhancing Multi-Character Portrait Animation with Expression-Augmented Diffusion Transformers},
  author={Wang, Qiang and Wang, Mengchao and Jiang, Fan and Fan, Yaqi and Qi, Yonggang and Xu, Mu},
  journal={arXiv preprint arXiv:2507.12956},
  year={2025}
}

@inproceedings{xu2025hunyuanportrait,
  title={Hunyuanportrait: Implicit condition control for enhanced portrait animation},
  author={Xu, Zunnan and Yu, Zhentao and Zhou, Zixiang and Zhou, Jun and Jin, Xiaoyu and Hong, Fa-Ting and Ji, Xiaozhong and Zhu, Junwei and Cai, Chengfei and Tang, Shiyu and others},
  booktitle={Proceedings of the Computer Vision and Pattern Recognition Conference},
  pages={15909--15919},
  year={2025}
}

@misc{drobyshev2024emoportraits,
      title={EMOPortraits: Emotion-enhanced Multimodal One-shot Head Avatars}, 
      author={Nikita Drobyshev and Antoni Bigata Casademunt and Konstantinos Vougioukas and Zoe Landgraf and Stavros Petridis and Maja Pantic},
      year={2024},
      eprint={2404.19110},
      archivePrefix={arXiv},
      primaryClass={cs.CV}
}

@article{qiu2025skyreels,
  title={Skyreels-a1: Expressive portrait animation in video diffusion transformers},
  author={Qiu, Di and Fei, Zhengcong and Wang, Rui and Bai, Jialin and Yu, Changqian and Fan, Mingyuan and Chen, Guibin and Wen, Xiang},
  journal={arXiv preprint arXiv:2502.10841},
  year={2025}
}

@inproceedings{wang2022pdfgc,
    title={Progressive Disentangled Representation Learning for Fine-Grained Controllable Talking Head Synthesis},
    author={Wang, Duomin and Deng, Yu and Yin, Zixin and Shum, Heung-Yeung and Wang, Baoyuan},
    booktitle={Proceedings of the IEEE/CVF Conference on Computer Vision and Pattern Recognition (CVPR)},
    year={2023}
}

@inproceedings{wang2021facevid2vid,
    title={One-Shot Free-View Neural Talking-Head Synthesis for Video Conferencing},
    author={Ting-Chun Wang and Arun Mallya and Ming-Yu Liu},
    booktitle={Proceedings of the IEEE Conference on Computer Vision and Pattern Recognition},
    year={2021}
}

@article{guo2024liveportrait,
  title   = {LivePortrait: Efficient Portrait Animation with Stitching and Retargeting Control},
  author  = {Guo, Jianzhu and Zhang, Dingyun and Liu, Xiaoqiang and Zhong, Zhizhou and Zhang, Yuan and Wan, Pengfei and Zhang, Di},
  journal = {arXiv preprint arXiv:2407.03168},
  year    = {2024}
}

@article{FLAME2017, 
  title = {Learning a model of facial shape and expression from {4D} scans}, 
  author = {Li, Tianye and Bolkart, Timo and Black, Michael. J. and Li, Hao and Romero, Javier}, 
  journal = {ACM Transactions on Graphics, (Proc. SIGGRAPH Asia)}, 
  volume = {36}, 
  number = {6}, 
  year = {2017}, 
  pages = {194:1--194:17},
  url = {https://doi.org/10.1145/3130800.3130813} 
}

@inproceedings{3dmm1999,
author = {Blanz, Volker and Vetter, Thomas},
title = {A morphable model for the synthesis of 3D faces},
year = {1999},
isbn = {0201485605},
publisher = {ACM Press/Addison-Wesley Publishing Co.},
address = {USA},
url = {https://doi.org/10.1145/311535.311556},
doi = {10.1145/311535.311556},
booktitle = {Proceedings of the 26th Annual Conference on Computer Graphics and Interactive Techniques},
pages = {187–194},
numpages = {8},
series = {SIGGRAPH '99}
}

@inproceedings{karras2019style,
  title={A style-based generator architecture for generative adversarial networks},
  author={Karras, Tero and Laine, Samuli and Aila, Timo},
  booktitle={Proceedings of the IEEE/CVF conference on computer vision and pattern recognition},
  pages={4401--4410},
  year={2019}
}

@InProceedings{Doukas_2021_ICCV,
    author    = {Doukas, Michail Christos and Zafeiriou, Stefanos and Sharmanska, Viktoriia},
    title     = {HeadGAN: One-Shot Neural Head Synthesis and Editing},
    booktitle = {Proceedings of the IEEE/CVF International Conference on Computer Vision (ICCV)},
    month     = {October},
    year      = {2021},
    pages     = {14398-14407}
}

@inproceedings{naruniec2020high,
  title={High-resolution neural face swapping for visual effects},
  author={Naruniec, Jacek and Helminger, Leonhard and Schroers, Christopher and Weber, Romann M},
  booktitle={Computer Graphics Forum},
  volume={39},
  number={4},
  pages={173--184},
  year={2020},
  organization={Wiley Online Library}
}

@inproceedings{ma2019real,
  title={Real-time hierarchical facial performance capture},
  author={Ma, Luming and Deng, Zhigang},
  booktitle={Proceedings of the ACM SIGGRAPH Symposium on Interactive 3D Graphics and Games},
  pages={1--10},
  year={2019}
}

@inproceedings{thies2016face2face,
  title={Face2face: Real-time face capture and reenactment of rgb videos},
  author={Thies, Justus and Zollhofer, Michael and Stamminger, Marc and Theobalt, Christian and Nie{\ss}ner, Matthias},
  booktitle={Proceedings of the IEEE conference on computer vision and pattern recognition},
  pages={2387--2395},
  year={2016}
}

@inproceedings{mildenhall2020nerf,
 title={NeRF: Representing Scenes as Neural Radiance Fields for View Synthesis},
 author={Ben Mildenhall and Pratul P. Srinivasan and Matthew Tancik and Jonathan T. Barron and Ravi Ramamoorthi and Ren Ng},
 year={2020},
 booktitle={ECCV},
}

@Article{kerbl3Dgaussians,
      author       = {Kerbl, Bernhard and Kopanas, Georgios and Leimk{\"u}hler, Thomas and Drettakis, George},
      title        = {3D Gaussian Splatting for Real-Time Radiance Field Rendering},
      journal      = {ACM Transactions on Graphics},
      number       = {4},
      volume       = {42},
      month        = {July},
      year         = {2023},
      url          = {https://repo-sam.inria.fr/fungraph/3d-gaussian-splatting/}
}

@inproceedings{ren2021pirenderer,
  title={Pirenderer: Controllable portrait image generation via semantic neural rendering},
  author={Ren, Yurui and Li, Ge and Chen, Yuanqi and Li, Thomas H and Liu, Shan},
  booktitle={Proceedings of the IEEE/CVF international conference on computer vision},
  pages={13759--13768},
  year={2021}
}

@article{thies2019deferred,
  title={Deferred neural rendering: Image synthesis using neural textures},
  author={Thies, Justus and Zollh{\"o}fer, Michael and Nie{\ss}ner, Matthias},
  journal={Acm Transactions on Graphics (TOG)},
  volume={38},
  number={4},
  pages={1--12},
  year={2019},
  publisher={ACM New York, NY, USA}
}

@inproceedings{deng2024portrait4d,
  title={Portrait4d: Learning one-shot 4d head avatar synthesis using synthetic data},
  author={Deng, Yu and Wang, Duomin and Ren, Xiaohang and Chen, Xingyu and Wang, Baoyuan},
  booktitle={Proceedings of the IEEE/CVF Conference on Computer Vision and Pattern Recognition},
  pages={7119--7130},
  year={2024}
}

@inproceedings{deng2024portrait4dv2,
  title={Portrait4d-v2: Pseudo multi-view data creates better 4d head synthesizer},
  author={Deng, Yu and Wang, Duomin and Wang, Baoyuan},
  booktitle={European Conference on Computer Vision},
  pages={316--333},
  year={2024},
  organization={Springer}
}

@inproceedings{
    chu2024gagavatar,
    title={Generalizable and Animatable Gaussian Head Avatar},
    author={Xuangeng Chu and Tatsuya Harada},
    booktitle={The Thirty-eighth Annual Conference on Neural Information Processing Systems},
    year={2024},
    url={https://openreview.net/forum?id=gVM2AZ5xA6}
}

@inproceedings{zeng2020realistic,
  title={Realistic face reenactment via self-supervised disentangling of identity and pose},
  author={Zeng, Xianfang and Pan, Yusu and Wang, Mengmeng and Zhang, Jiangning and Liu, Yong},
  booktitle={Proceedings of the AAAI conference on artificial intelligence},
  volume={34},
  number={07},
  pages={12757--12764},
  year={2020}
}

@inproceedings{siarohin2019animating,
  title={Animating arbitrary objects via deep motion transfer},
  author={Siarohin, Aliaksandr and Lathuili{\`e}re, St{\'e}phane and Tulyakov, Sergey and Ricci, Elisa and Sebe, Nicu},
  booktitle={Proceedings of the IEEE/CVF conference on computer vision and pattern recognition},
  pages={2377--2386},
  year={2019}
}

@article{siarohin2019first,
  title={First order motion model for image animation},
  author={Siarohin, Aliaksandr and Lathuili{\`e}re, St{\'e}phane and Tulyakov, Sergey and Ricci, Elisa and Sebe, Nicu},
  journal={Advances in neural information processing systems},
  volume={32},
  year={2019}
}

@inproceedings{siarohin2021motion,
  title={Motion representations for articulated animation},
  author={Siarohin, Aliaksandr and Woodford, Oliver J and Ren, Jian and Chai, Menglei and Tulyakov, Sergey},
  booktitle={Proceedings of the IEEE/CVF conference on computer vision and pattern recognition},
  pages={13653--13662},
  year={2021}
}

@inproceedings{zakharov2019few,
  title={Few-shot adversarial learning of realistic neural talking head models},
  author={Zakharov, Egor and Shysheya, Aliaksandra and Burkov, Egor and Lempitsky, Victor},
  booktitle={Proceedings of the IEEE/CVF international conference on computer vision},
  pages={9459--9468},
  year={2019}
}

@inproceedings{burkov2020neural,
  title={Neural head reenactment with latent pose descriptors},
  author={Burkov, Egor and Pasechnik, Igor and Grigorev, Artur and Lempitsky, Victor},
  booktitle={Proceedings of the IEEE/CVF conference on computer vision and pattern recognition},
  pages={13786--13795},
  year={2020}
}

@inproceedings{wiles2018x2face,
  title={X2face: A network for controlling face generation using images, audio, and pose codes},
  author={Wiles, Olivia and Koepke, A and Zisserman, Andrew},
  booktitle={Proceedings of the European conference on computer vision (ECCV)},
  pages={670--686},
  year={2018}
}

@inproceedings{zakharov2020fast,
  title={Fast bi-layer neural synthesis of one-shot realistic head avatars},
  author={Zakharov, Egor and Ivakhnenko, Aleksei and Shysheya, Aliaksandra and Lempitsky, Victor},
  booktitle={European Conference on Computer Vision},
  pages={524--540},
  year={2020},
  organization={Springer}
}

@inproceedings{yang2022face2face,
  title={Face2face $\rho$: Real-time high-resolution one-shot face reenactment},
  author={Yang, Kewei and Chen, Kang and Guo, Daoliang and Zhang, Song-Hai and Guo, Yuan-Chen and Zhang, Weidong},
  booktitle={European conference on computer vision},
  pages={55--71},
  year={2022},
  organization={Springer}
}

@inproceedings{hong2022depth,
  title={Depth-aware generative adversarial network for talking head video generation},
  author={Hong, Fa-Ting and Zhang, Longhao and Shen, Li and Xu, Dan},
  booktitle={Proceedings of the IEEE/CVF conference on computer vision and pattern recognition},
  pages={3397--3406},
  year={2022}
}

@inproceedings{liu2025avatarartist,
  title={Avatarartist: Open-domain 4d avatarization},
  author={Liu, Hongyu and Wang, Xuan and Wan, Ziyu and Ma, Yue and Chen, Jingye and Fan, Yanbo and Shen, Yujun and Song, Yibing and Chen, Qifeng},
  booktitle={Proceedings of the Computer Vision and Pattern Recognition Conference},
  pages={10758--10769},
  year={2025}
}

@inproceedings{smirk2024,
  title={3d facial expressions through analysis-by-neural-synthesis},
  author={Retsinas, George and Filntisis, Panagiotis P and Danecek, Radek and Abrevaya, Victoria F and Roussos, Anastasios and Bolkart, Timo and Maragos, Petros},
  booktitle={Proceedings of the IEEE/CVF Conference on Computer Vision and Pattern Recognition},
  pages={2490--2501},
  year={2024}
}

@article{wei2024aniportrait,
  title={Aniportrait: Audio-driven synthesis of photorealistic portrait animation},
  author={Wei, Huawei and Yang, Zejun and Wang, Zhisheng},
  journal={arXiv preprint arXiv:2403.17694},
  year={2024}
}

@inproceedings{guo2025high,
  title={High-Fidelity Relightable Monocular Portrait Animation with Lighting-Controllable Video Diffusion Model},
  author={Guo, Mingtao and Xing, Guanyu and Liu, Yanli},
  booktitle={Proceedings of the Computer Vision and Pattern Recognition Conference},
  pages={228--238},
  year={2025}
}

@inproceedings{vace,
    title = {VACE: All-in-One Video Creation and Editing},
    author = {Jiang, Zeyinzi and Han, Zhen and Mao, Chaojie and Zhang, Jingfeng and Pan, Yulin and Liu, Yu},
    booktitle = {Proceedings of the IEEE/CVF International Conference on Computer Vision},
    pages = {17191-17202},
    year = {2025}
}

@article{tan2025fixtalk,
  title={FixTalk: Taming Identity Leakage for High-Quality Talking Head Generation in Extreme Cases},
  author={Tan, Shuai and Gong, Bill and Ji, Bin and Pan, Ye},
  journal={arXiv preprint arXiv:2507.01390},
  year={2025}
}

@inproceedings{tan2024edtalk,
  title={Edtalk: Efficient disentanglement for emotional talking head synthesis},
  author={Tan, Shuai and Ji, Bin and Bi, Mengxiao and Pan, Ye},
  booktitle={European Conference on Computer Vision},
  pages={398--416},
  year={2024},
  organization={Springer}
}

@inproceedings{park2019semantic,
  title={Semantic image synthesis with spatially-adaptive normalization},
  author={Park, Taesung and Liu, Ming-Yu and Wang, Ting-Chun and Zhu, Jun-Yan},
  booktitle={Proceedings of the IEEE/CVF conference on computer vision and pattern recognition},
  pages={2337--2346},
  year={2019}
}

@article{oquab2023dinov2,
  title={Dinov2: Learning robust visual features without supervision},
  author={Oquab, Maxime and Darcet, Timoth{\'e}e and Moutakanni, Th{\'e}o and Vo, Huy and Szafraniec, Marc and Khalidov, Vasil and Fernandez, Pierre and Haziza, Daniel and Massa, Francisco and El-Nouby, Alaaeldin and others},
  journal={arXiv preprint arXiv:2304.07193},
  year={2023}
}

@inproceedings{zielonka2022towards,
  title={Towards metrical reconstruction of human faces},
  author={Zielonka, Wojciech and Bolkart, Timo and Thies, Justus},
  booktitle={European conference on computer vision},
  pages={250--269},
  year={2022},
  organization={Springer}
}

@inproceedings{feng2018wing,
  title={Wing loss for robust facial landmark localisation with convolutional neural networks},
  author={Feng, Zhen-Hua and Kittler, Josef and Awais, Muhammad and Huber, Patrik and Wu, Xiao-Jun},
  booktitle={Proceedings of the IEEE conference on computer vision and pattern recognition},
  pages={2235--2245},
  year={2018}
}

@article{wan2025wan,
  title={Wan: Open and advanced large-scale video generative models},
  author={Wan, Team and Wang, Ang and Ai, Baole and Wen, Bin and Mao, Chaojie and Xie, Chen-Wei and Chen, Di and Yu, Feiwu and Zhao, Haiming and Yang, Jianxiao and others},
  journal={arXiv preprint arXiv:2503.20314},
  year={2025}
}

@inproceedings{deng2019arcface,
  title={Arcface: Additive angular margin loss for deep face recognition},
  author={Deng, Jiankang and Guo, Jia and Xue, Niannan and Zafeiriou, Stefanos},
  booktitle={Proceedings of the IEEE/CVF conference on computer vision and pattern recognition},
  pages={4690--4699},
  year={2019}
}

@article{wang2004image,
  title={Image quality assessment: from error visibility to structural similarity},
  author={Wang, Zhou and Bovik, Alan C and Sheikh, Hamid R and Simoncelli, Eero P},
  journal={IEEE transactions on image processing},
  volume={13},
  number={4},
  pages={600--612},
  year={2004},
  publisher={IEEE}
}

@inproceedings{zhang2018unreasonable,
  title={The unreasonable effectiveness of deep features as a perceptual metric},
  author={Zhang, Richard and Isola, Phillip and Efros, Alexei A and Shechtman, Eli and Wang, Oliver},
  booktitle={Proceedings of the IEEE conference on computer vision and pattern recognition},
  pages={586--595},
  year={2018}
}

@article{heusel2017gans,
  title={Gans trained by a two time-scale update rule converge to a local nash equilibrium},
  author={Heusel, Martin and Ramsauer, Hubert and Unterthiner, Thomas and Nessler, Bernhard and Hochreiter, Sepp},
  journal={Advances in neural information processing systems},
  volume={30},
  year={2017}
}

@article{unterthiner2019fvd,
  title={FVD: A new metric for video generation},
  author={Unterthiner, Thomas and Van Steenkiste, Sjoerd and Kurach, Karol and Marinier, Rapha{\"e}l and Michalski, Marcin and Gelly, Sylvain},
  year={2019}
}

@inproceedings{han2024face,
  title={Face-adapter for pre-trained diffusion models with fine-grained id and attribute control},
  author={Han, Yue and Zhu, Junwei and He, Keke and Chen, Xu and Ge, Yanhao and Li, Wei and Li, Xiangtai and Zhang, Jiangning and Wang, Chengjie and Liu, Yong},
  booktitle={European Conference on Computer Vision},
  pages={20--36},
  year={2024},
  organization={Springer}
}

@inproceedings{woo2023convnext,
  title={Convnext v2: Co-designing and scaling convnets with masked autoencoders},
  author={Woo, Sanghyun and Debnath, Shoubhik and Hu, Ronghang and Chen, Xinlei and Liu, Zhuang and Kweon, In So and Xie, Saining},
  booktitle={Proceedings of the IEEE/CVF conference on computer vision and pattern recognition},
  pages={16133--16142},
  year={2023}
}

@InProceedings{xie2022vfhq,
      author = {Liangbin Xie and Xintao Wang and Honglun Zhang and Chao Dong and Ying Shan},
      title = {VFHQ: A High-Quality Dataset and Benchmark for Video Face Super-Resolution},
      booktitle={The IEEE Conference on Computer Vision and Pattern Recognition Workshops (CVPRW)},
      year = {2022}
  }

@inproceedings{kaisiyuan2020mead,
 author = {Wang, Kaisiyuan and Wu, Qianyi and Song, Linsen and Yang, Zhuoqian and Wu, Wayne and Qian, Chen and He, Ran and Qiao, Yu and Loy, Chen Change},
 title = {MEAD: A Large-scale Audio-visual Dataset for Emotional Talking-face Generation},
 booktitle = {ECCV},
 month = Augest,
 year = {2020}
}

@article{kirschstein2023nersemble,
    author = {Kirschstein, Tobias and Qian, Shenhan and Giebenhain, Simon and Walter, Tim and Nie\ss{}ner, Matthias},
    title = {NeRSemble: Multi-View Radiance Field Reconstruction of Human Heads},
    year = {2023},
    issue_date = {August 2023},
    publisher = {Association for Computing Machinery},
    address = {New York, NY, USA},
    volume = {42},
    number = {4},
    issn = {0730-0301},
    url = {https://doi.org/10.1145/3592455},
    doi = {10.1145/3592455},
    journal = {ACM Trans. Graph.},
    month = {jul},
    articleno = {161},
    numpages = {14},
}

@inproceedings{kim2022adaface,
  title={AdaFace: Quality Adaptive Margin for Face Recognition},
  author={Kim, Minchul and Jain, Anil K and Liu, Xiaoming},
  booktitle={Proceedings of the IEEE/CVF Conference on Computer Vision and Pattern Recognition},
  year={2022}
}

@inproceedings{Zhang2020ETHXGaze,
  author    = {Xucong Zhang and Seonwook Park and Thabo Beeler and Derek Bradley and Siyu Tang and Otmar Hilliges},
  title     = {ETH-XGaze: A Large Scale Dataset for Gaze Estimation under Extreme Head Pose and Gaze Variation},
  year      = {2020},
  booktitle = {European Conference on Computer Vision (ECCV)}
}

@article{kingma2014adam,
  title={Adam: A method for stochastic optimization},
  author={Kingma, Diederik P},
  journal={arXiv preprint arXiv:1412.6980},
  year={2014}
}

@misc{runway,
  title        = {Creating with act-two},
  author       = {Runway},
  year         = {2025},
  howpublished = {https://help.runwayml.com/hc/en-us/articles/42311337895827-Creating-with-Act-Two},
}

@article{krizhevsky2012imagenet,
  title={Imagenet classification with deep convolutional neural networks},
  author={Krizhevsky, Alex and Sutskever, Ilya and Hinton, Geoffrey E},
  journal={Advances in neural information processing systems},
  volume={25},
  year={2012}
}

@inproceedings{abdelrahman2023l2cs,
  title={L2cs-net: Fine-grained gaze estimation in unconstrained environments},
  author={Abdelrahman, Ahmed A and Hempel, Thorsten and Khalifa, Aly and Al-Hamadi, Ayoub and Dinges, Laslo},
  booktitle={2023 8th International Conference on Frontiers of Signal Processing (ICFSP)},
  pages={98--102},
  year={2023},
  organization={IEEE}
}

@inproceedings{szegedy2016rethinking,
  title={Rethinking the inception architecture for computer vision},
  author={Szegedy, Christian and Vanhoucke, Vincent and Ioffe, Sergey and Shlens, Jon and Wojna, Zbigniew},
  booktitle={Proceedings of the IEEE conference on computer vision and pattern recognition},
  pages={2818--2826},
  year={2016}
}

@inproceedings{carreira2017quo,
  title={Quo vadis, action recognition? a new model and the kinetics dataset},
  author={Carreira, Joao and Zisserman, Andrew},
  booktitle={proceedings of the IEEE Conference on Computer Vision and Pattern Recognition},
  pages={6299--6308},
  year={2017}
}

@misc{metahuman,
  title        = {Metahuman Creator},
  author       = {Epic Games},
  year         = {2021},
  howpublished = {https://www.unrealengine.com/en-US/digital-humans},
}

@inproceedings{
  xu2022vitpose,
  title={Vi{TP}ose: Simple Vision Transformer Baselines for Human Pose Estimation},
  author={Yufei Xu and Jing Zhang and Qiming Zhang and Dacheng Tao},
  booktitle={Advances in Neural Information Processing Systems},
  year={2022},
}

@inproceedings{
    wang2022latent,
    title={Latent Image Animator: Learning to Animate Images via Latent Space Navigation},
    author={Yaohui Wang and Di Yang and Francois Bremond and Antitza Dantcheva},
    booktitle={International Conference on Learning Representations},
    year={2022}
}

@article{wang2025lia,
  title={LIA-X: Interpretable Latent Portrait Animator},
  author={Wang, Yaohui and Yang, Di and Chen, Xinyuan and Bremond, Francois and Qiao, Yu and Dantcheva, Antitza},
  journal={arXiv preprint arXiv:2508.09959},
  year={2025}
}

\clearpage
\appendix

\section*{Table of Contents}
We first provide a brief overview of our supplementary material. This supplementary material consists of the following sections and contents:
\begin{itemize}
\item \cref{sec:LLM}: LLM usage claim.
\item \cref{sec:pixel3dmm}: The detailed process of our used 3DMM-based face tracking method Pixel3DMM~\cite{giebenhain2025pixel3dmm}.
\item \cref{sec:loss}: Definitions of all training loss terms used in our model.
\item \cref{sec:mask}: Describes the facial and non-facial masks calculation process of each frame.
\item \cref{sec:inference}: Provides the inference process of portrait animation task.
\item \cref{sec:keypoints}: Shows our keypoints selection strategy.
\item \cref{sec:training}: Contains the specific training settings to train our model.
\item \cref{sec:metrics}: The detailed definition of each evaluation metric.
\item \cref{sec:benchmark}: Construction pipeline of our test benchmark.
\item \cref{sec:modification}: How we modify four diffusion-based methods to support the portrait video editing task.
\item \cref{sec:results}: Provides more generated results of our PerformRecast, including both portrait video expression editing and portrait animation.
\item \cref{sec:limitation}: Discusses the limitations of our method and future plans.
\item \cref{sec:ethics}: Ethics statement of our method to avoid malicious use.
\end{itemize}

\section{LLM Use Claim} \label{sec:LLM}
We employ a large language model (LLM) to assist with the language polishing and revision of certain sections of our paper, including the supplementary material. The LLM is used solely to enhance grammar, clarity, and overall readability by rephrasing sentences, correcting linguistic errors, and ensuring stylistic consistency. All authors have carefully reviewed and approved the final manuscript and take full responsibility for its content.

\section{3DMM-based Face Tracking} \label{sec:pixel3dmm}
To obtain temporally continuous FLAME~\cite{FLAME2017} model reconstruction results from input portrait videos, We adopt a recently-proposed 3D face tracking method, Pixel3DMM~\cite{giebenhain2025pixel3dmm} to predict FLAME parameters of each frame from input portrait videos. Pixel3DMM firstly trains two expert networks: $\mathcal{N}$ and $\mathcal{U}$, which are built on the top of the pretrained DINOv2~\cite{oquab2023dinov2} backbone to predict surface normal $\mathcal{N}(I)$ and UV-space coordinate $\mathcal{U}(I)$ given a portrait image $I$.

Then, it optimizes for FLAME parameters~\cite{FLAME2017}, including face identity $\beta \in \mathbb{R}^{300}$, expression $\psi \in \mathbb{R}^{100}$, head pose $\theta \in \mathbb{R}^{3 * 4 + 3 = 15}$ and other camera parameters. The head pose $\theta$ contains four 3D rotation vectors for four joints: $\theta_\text{neck}, \theta_\text{jaw}, \theta_\text{left-eyeball}$, $\theta_\text{right-eyeball}$ and one global rotation $\theta_\text{head}$ in axis-angle. Specifically, Pixel3DMM directly uses MICA's~\cite{zielonka2022towards} identity prediction as $\beta$. The remaining parameters are optimized via minimizing a 2D vertex loss and a normal rendering loss between the projection of current estimated FLAME model and predicted UV-space coordinate $\mathcal{U}(I)$ as well as surface normal $\mathcal{N}(I)$.

For monocular video tracking, Pixel3DMM freezes $\mathbf{z}_\text{id}$ using the average result of MICA's~\cite{zielonka2022towards} identity predictions across all frames. Then, it sequentially optimize for the remaining parameters for each frame. Finally, it adds a smoothness term to ensure smoothness across all frames.

As a result, Pixel3DMM is capable of reconstructing temporally continuous FLAME parameters and fixed face identity of each input portrait video.

\begin{figure}[t!]
  \centering
  \includegraphics[width=\columnwidth]{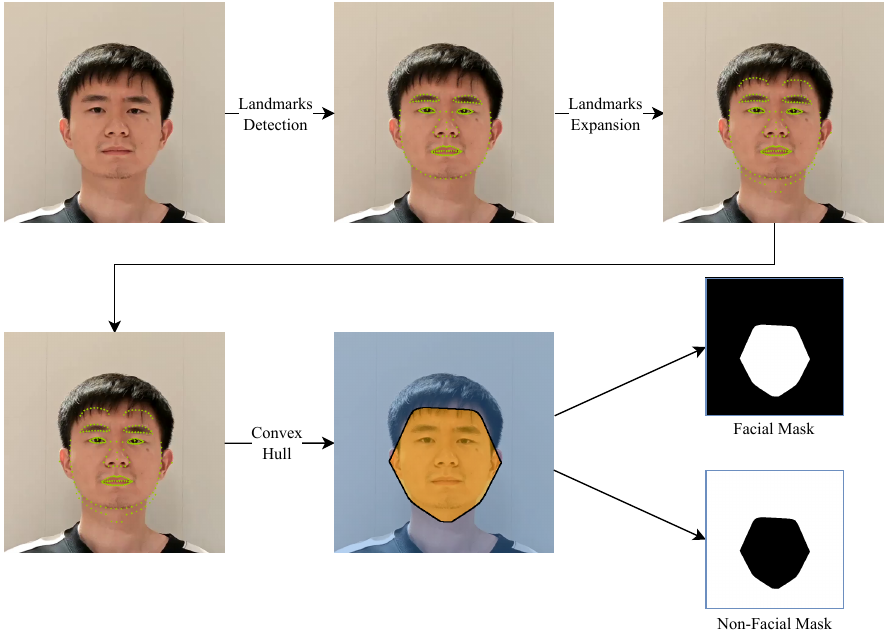}
  \caption{The facial mask calculation process of each frame in training dataset.}
  \label{fig:teacher-student}
  \vspace{-5mm}
\end{figure}

\section{Definitions of Training Loss Terms} \label{sec:loss}
We utilize $\mathcal{L}_\text{animate}$ which is described in the main manuscript to train our teacher and student models. $\mathcal{L}_\text{animate}$ is formulated as:
\begin{equation}
    \mathcal{L}_\text{animate} = \mathcal{L}_\text{FLAME} + \mathcal{L}_{P,\text{cascade}} + \mathcal{L}_{1,\text{cascade}} +\mathcal{L}_{G, \text{cascade}} + \mathcal{L}_\text{faceid},
\label{equ:animate-suppl}
\end{equation}
To calculate the difference between the reconstructed frame $\hat{I}_d$ and driving frame $I_d$, we utilize three commonly-used loss term: the perceptual loss, $L_1$ loss and GAN-loss. To further improve the texture quality, the perceptual loss, $L_1$ loss and GAN loss are applied on both global region and local regions of face and lip, which are denoted as a cascaded perceptual loss $\mathcal{L}_{P,\text{cascade}}$, a cascaded $L_1$ loss $\mathcal{L}_{1, \text{cascade}}$ and a cascaded GAN loss $\mathcal{L}_{G, \text{cascade}}$. $\mathcal{L}_{G, \text{cascade}}$ consists of $\mathcal{L}_{\text{GAN}, \text{global}}$, $\mathcal{L}_{\text{GAN}, \text{face}}$ and $\mathcal{L}_{\text{GAN}, \text{lip}}$, which depend on the corresponding discriminators $\mathcal{D}_\text{global}$, $\mathcal{D}_\text{face}$ and $\mathcal{D}_\text{lip}$ training from scratch. The face and lip regions are defined using the 2D semantic facial landmarks which are extracted by a pre-trained landmark detector in LivePortrait~\cite{guo2024liveportrait}. And the face-id~\cite{deng2019arcface} loss is used to preserve the identity of source image $I_s$.

\section{Facial Mask Calculation} \label{sec:mask}
As shown in~\cref{fig:teacher-student}, to obtain masks of facial and non-facial regions, we also utilize the pre-trained 2D facial landmark detector in LivePortrait~\cite{guo2024liveportrait} to extract 203 landmarks of each frame from our dataset. Then, we expand the detected 2D facial landmarks of source frame $I_s$ outward and compute their convex hull as the facial region, while the remaining area in $I_s$ is regarded as the non-facial region. 

\section{Inference Process of Portrait Animation} \label{sec:inference}
In the inference phase of portrait animation task, we first extract the appearance feature volume $f_s = \mathcal{F}(I_s)$ from the source image $I_s$. Given a driving video sequence $\{I_{d,i}| i = 0,1,...,N-1\}$, the source and driving explicit keypoints are transformed as follows:
\begin{equation}
    \left\{
        \begin{array}{lr}
        x_{s} = s_s \cdot \big((x_{c,s} +\delta_{s}) R_s \big) + t_{s}, &  \\
        x_{d,i} = s_{d,i} \cdot \big((x_{c,s} + \delta_{d,i}) R_{d,i} \big) + t_{d,i}, &
        \end{array}
    \right.
    \label{equ:portrait_animation}
\end{equation}
which utilizes the same formula as training stage.

\begin{figure}[t!]
  \centering
  \includegraphics[width=\columnwidth]{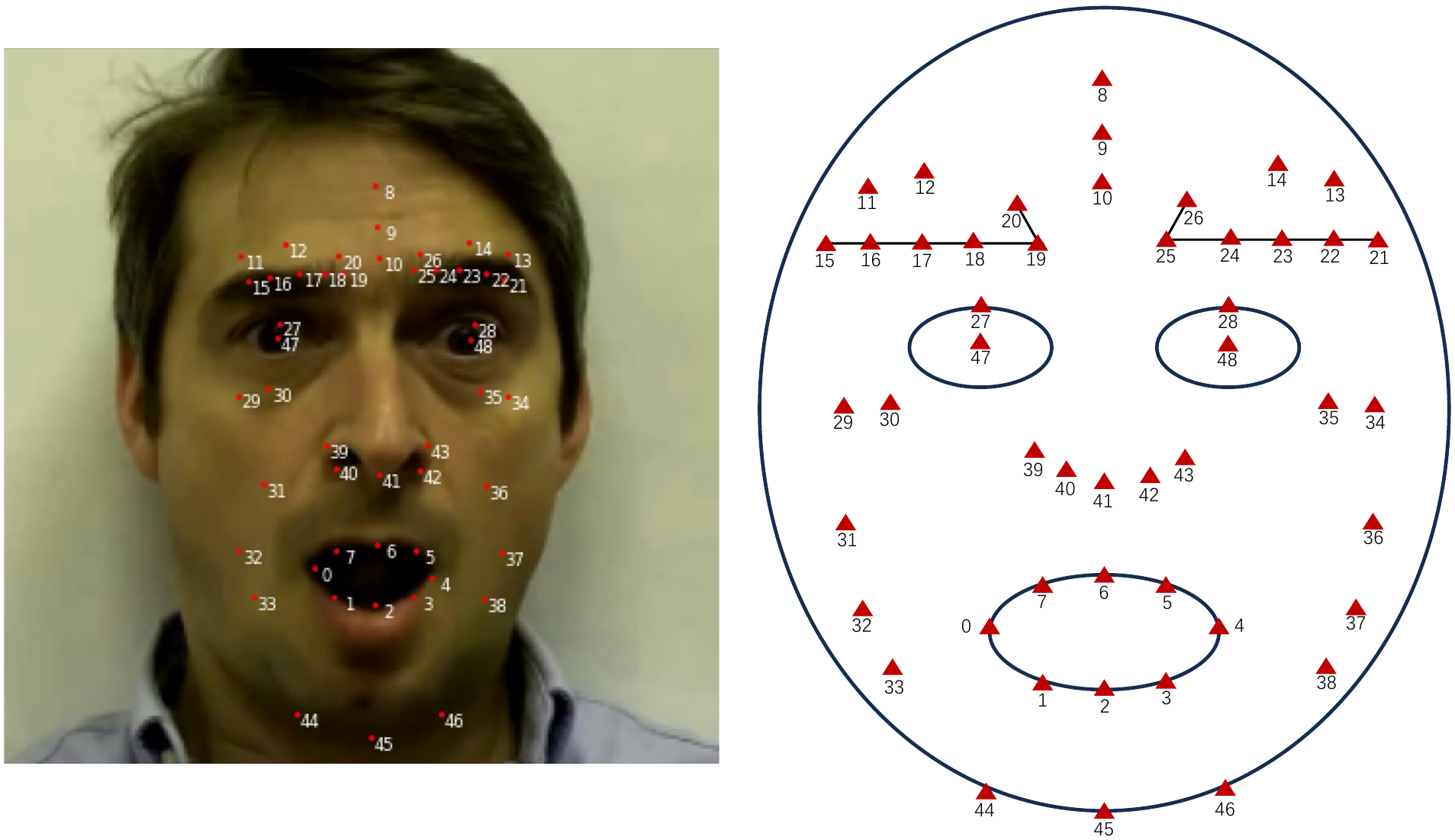}
  \caption{The specific location of each keypoint used in our method. Please zoom in for better inspection.}
  \label{fig:keypoints}
  \vspace{-5mm}
\end{figure}

\section{Keypoints Selection} \label{sec:keypoints}
We select $K = 49$ keypoints from the reconstructed FLAME face mesh in total to supervise our motion extractor. ~\cref{fig:keypoints} shows the specific location of each keypoint. We select as few keypoints as possible, covering important facial regions such as the forehead, eyebrows, eye sockets, eyeballs, nose, lip, and jaw.

\section{Training Settings} \label{sec:training}
We train our model from scratch using 128 NVIDIA H20 GPUs for approximately one week with a batch size of 8 per GPU. We adopt the Adam~\cite{kingma2014adam} optimizer with different learning rates for different modules. Specifically, the appearance feature extractor is trained with a learning rate of $5\times10^{-5}$, while the motion extractor, warping module, and decoder are assigned a higher learning rate of $1.2\times10^{-4}$. To further stabilize adversarial training, we set the learning rates of the image, face, and lip discriminators to $1\times10^{-4}$, $2.5\times10^{-5}$, and $1.5\times10^{-5}$, respectively. To improve the robustness of training process, we further add random gaussian noise with small variance on extracted keypoints $x_s$ and $x_d$, but not during inference stage.

\section{Evaluation Metrics Details} \label{sec:metrics}
\paragraph{LPIPS.}
For potrait video expression editing and self-reenactment, we calculate the perceptual similarity metric LPIPS~\cite{zhang2018unreasonable} based on AlexNet~\cite{krizhevsky2012imagenet} between the animated images and ground truth images.

\vspace{-3mm}
\paragraph{Fréchet Inception Distance.} For portrait video expression editing and self-reenactment, FID compares the distribution of generated images with the distribution of ground truth images. The formula for FID is defined as:
\begin{equation}
\text{FID}=||\mu_g-\mu_r||^2 + \text{Tr}(\Sigma_g + \Sigma_r - 2(\Sigma_r\Sigma_g)^{1/2}),
\label{equ:fid}
\end{equation}
where $g$ and $r$ denote the features of the generated image and ground truth images, which is extracted by Inception-v3 model~\cite{szegedy2016rethinking}. $\mu$ and $\Sigma$ denote the mean and covariance matrices of each image set. A lower FID indicates better generation quality.

\vspace{-3mm}
\paragraph{Fréchet Video Distance.}
For portrait video expression editing and self-reenactment, FVD compares the distribution of generated videos with the distribution of ground truth videos. The formula for FVD is similar to FID, which is defined as:
\begin{equation}
\text{FVD}=||\mu_g-\mu_r||^2 + \text{Tr}(\Sigma_g + \Sigma_r - 2(\Sigma_r\Sigma_g)^{1/2}),
\label{equ:fvd}
\end{equation}
where $g$ and $r$ denote the features of the generated videos and ground truth videos, which is extract by the pre-trained Inflated 3D ConvNet~\cite{carreira2017quo}. $\mu$ and $\Sigma$ denote the mean and covariance matrices of each video set. A lower FVD indicates better generation quality.

\vspace{-3mm}
\paragraph{Cosine SIMilarity of identity features.} We utilize CSIM  to measure the identity preservation between two images, through the cosine similarity of two embeddings from a recently proposed pretrained face recognition network AdaFace~\cite{kim2022adaface}. For portrait video expression editing and self-reenactment, the CSIM is calculated between the animated image and ground truth image. For cross-reenactment, the CSIM is calculated between the animated and the source images.

\vspace{-3mm}
\paragraph{Average Expression Distance.} AED is the mean $L_1$ distance of the expression parameters between the edited and driving images in expression editing task as well as the animated and driving images in portrait animation task. These parameters, which include expression coefficient, eyelid and jaw pose parameters, are extracted by the state-of-the-art 3D face reconstruction method SMIRK~\cite{smirk2024}.

\vspace{-3mm}
\paragraph{Average Pose Distance.} APD is the mean $L_1$ distance of the pose parameters between the edited and source images in expression editing task as well as the animated and driving images in portrait animation task. The pose parameters are also extract by SMIRK~\cite{smirk2024}.

\vspace{-3mm}
\paragraph{Mean Angular Error.} The mean angular error is used to measure the eyeball direction error between the edited and driving images in expression editing task as well as the animated and driving images in portrait animation task. It is adopted as: 
$
    \textrm{MAE}(I_{g}, I_{d}) = \textrm{arccos}(\frac{\textbf{b}_g \cdot \textbf{b}_d}{\Vert \textbf{b}_g \Vert \cdot \Vert \textbf{b}_d \Vert}),
$ where $\textbf{b}_g$ and $\textbf{b}_d$ are the eyeball direction vectors of the generated image $I_{g}$ (including the edited image and animated image) and the driving image $I_{d}$ respectively. Both of them are predicted by a pre-trained eyeball direction prediction network~\cite{abdelrahman2023l2cs}.

\begin{figure}[t!]
  \centering
  \includegraphics[width=\columnwidth]{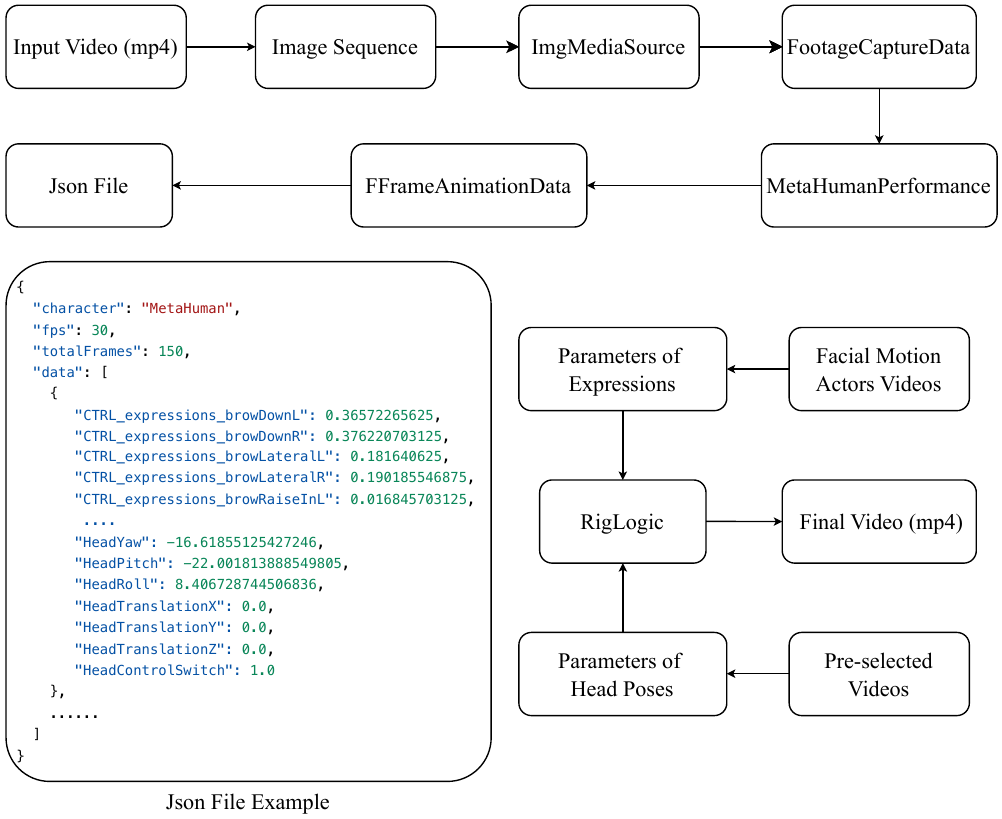}
  \caption{Construction pipeline of our proposed test benchmark.}
  \label{fig:benchmark}
  \vspace{-5mm}
\end{figure}

\section{Construction of Our Test Benchmark} \label{sec:benchmark}
~\cref{fig:benchmark} visualizes the construction pipeline of our proposed test benchmark. Given a input video, our pipeline first utilize MetaHuman~\cite{metahuman} to extract the expression and head pose parameters of each frame. The detailed information of this process are shown at the top of ~\cref{fig:benchmark}. The extracted parameters of each frame are saved in a json file. Among them, the keys of parameters related to facial expressions start with ``CTRL\_expressions". The three keys ``HeadYaw'', ``HeadPitch'' and ``HeadRoll'' describe the head pose rotation. Then, we combine the parameters with keys starting with ``CTRL\_expressions" in the json files extracted from facial motion actors and ``HeadYaw'', ``HeadPitch'', ``HeadRoll'' in the json files extracted from our pre-selected videos containing large head pose rotation to RigLogic system to drive the ditigal human in MetaHuman and create final videos. For enhancement mode, all parameters with keys starting with ``CTRL\_expressions" are set to zero.

The resolution of original videos synthesized from MetaHuman are set to $2560 \times 1440$, which is the default setting. Each video contains 150 frames and is recorded with 30 frames per second (FPS). We crop all the videos into squares to maintain the face at the center and resize them to the resolution of $512 \times 512$ for further training. 

\section{Modification of Diffusion-based Methods} \label{sec:modification}
We modify several diffusion-based portrait animation methods to make them support the task of editing the facial expression of source video according to the driving video. All these four methods leverage large-scale pre-trained video diffusion models to animate the input static portrait image from the driving video. However, our portrait video expression editing task needs to utilize the $i$-th frame $I_{d,i}$ in driving video to edit the expression of $i$-th frame $I_{s,i}$ in source video. Therefore, we expand each frame of the driving video into a short static video clip, which is then used to animate the $i$-th frame of the source video, thus conforming to the video input formula required by video diffusion models. For the source and driving video of N frames, we repeat this animation process for N times, and concatenate N animated images to form the edited video. 

To realize expression editing instead of portrait animation, the key idea is to combine the facial expression of driving frame with the head pose of source frame, and use this combined signal to animate the source frame. We then describe the detailed modification of each method.

\begin{figure}[t!]
  \centering
  \includegraphics[width=\columnwidth]{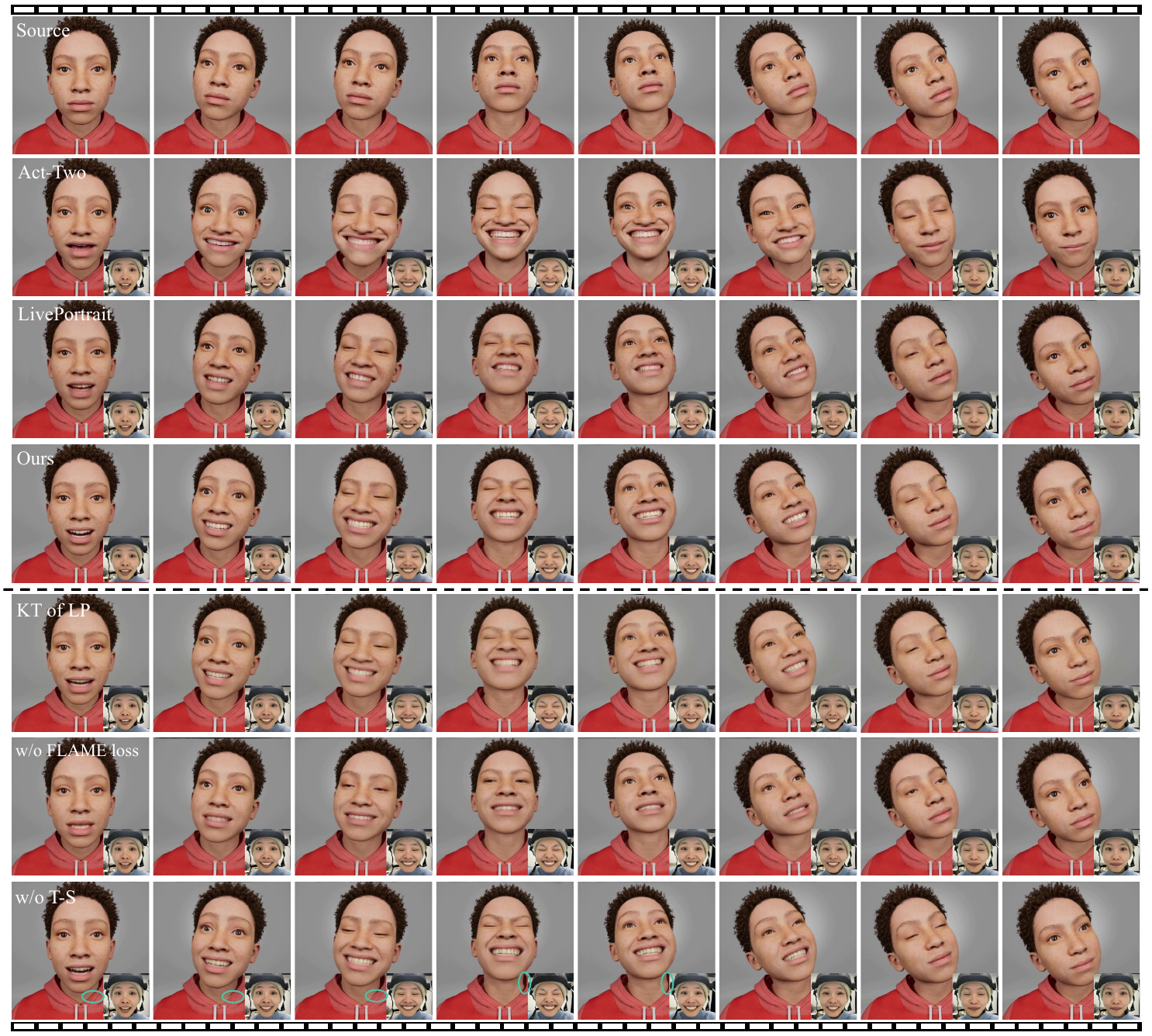}
  \caption{Qualitative comparison of portrait video expression editing on enhancement mode. The top of the figure shows editing results of different methods. The bottom presents our ablation studies and analysis. The bottom-right insets exhibit driving frames. The light green circles highlight the misalignment between the facial and non-facial regions. Please zoom in for better inspection.}
  \label{fig:enhancement}
  \vspace{-6mm}
\end{figure}

\begin{figure*}[t!]
  \centering
  \includegraphics[width=\linewidth]{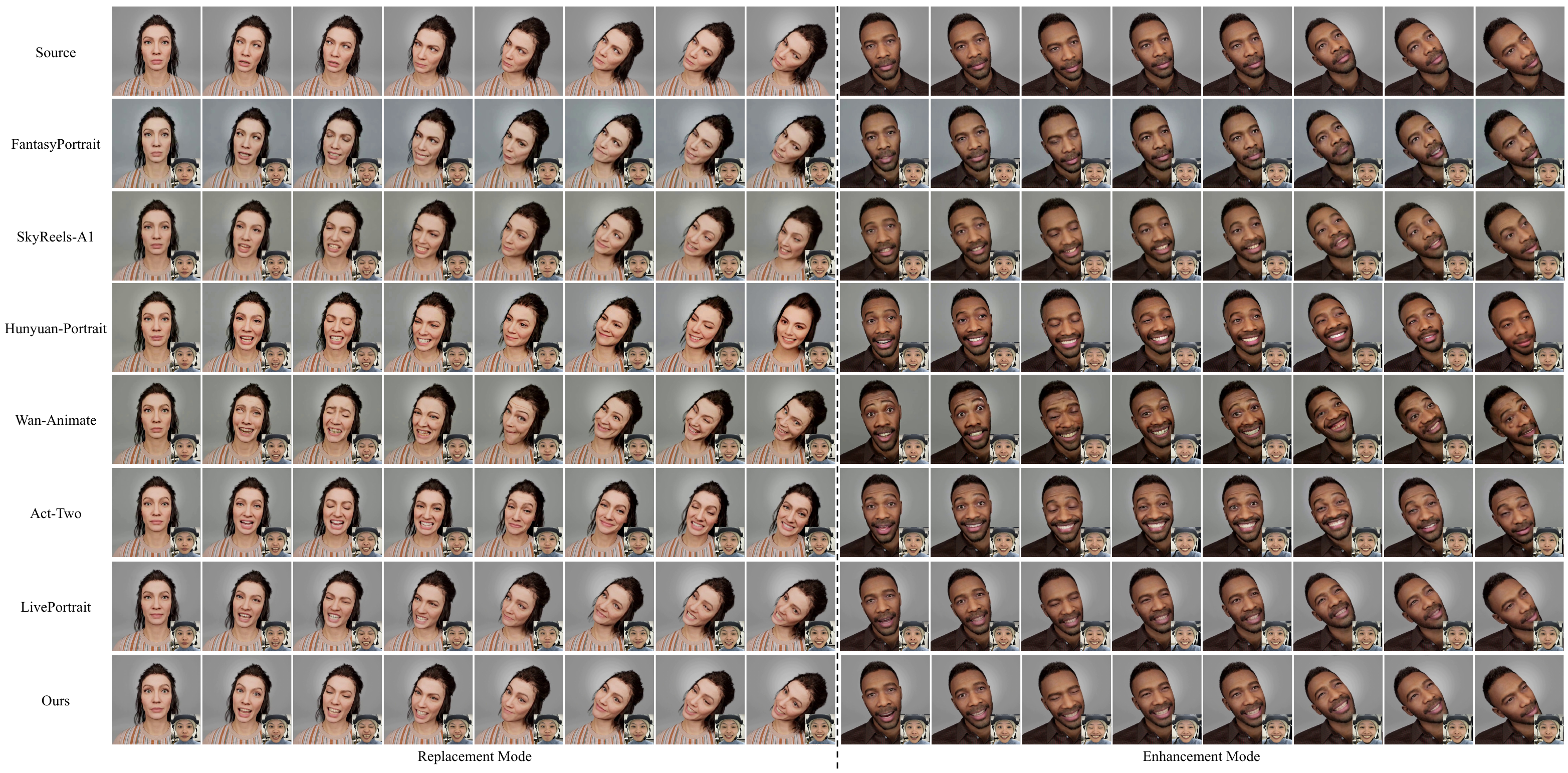}
  \caption{Full qualitative comparison with all the methods mentioned in the main manuscript on our proposed test benchmark. The bottom-right insets exhibit driving frames. Please zoom in for better inspection.}
  \label{fig:expression}
\end{figure*}

\vspace{-3mm}
\paragraph{SkyReels-A1.}
SkyReels-A1~\cite{qiu2025skyreels} utilizes SMIRK~\cite{smirk2024} to extract FLAME~\cite{FLAME2017} parameters of each frame in driving video. In our task, we replace the head pose parameters in FLAME model of driving frame with that of source frame to animate the source frame.

\vspace{-3mm}
\paragraph{Hunyuan-Portrait.}
Hunyuan-Portrait~\cite{xu2025hunyuanportrait} utilizes pre-trained motion encoder MegaPortraits~\cite{Drobyshev22MP} to extract facial motion representations of driving video. Specifically, these representations consist of the explicit head rotations $R$, translations $t$, and the latent expression descriptors $z$. Therefore, we replace the head rotations $R$ and translations $t$ of driving frame with those of source frame to animate the source frame.

\vspace{-3mm}
\paragraph{FantasyPortrait.}
FantasyPortrait employs a pre-trained implicit expression motion extractor PD-FGC~\cite{wang2022pdfgc} to encode the driving frame into latent features. These latent features include lip motion $e_{lip}$, eye gaze and blink $e_{eye}$, head pose $e_{head}$ and emotional expression $e_{emo}$. And we replace the head pose parameters $e_{head}$ of driving frame with that of source frame to animate the source frame.

\vspace{-3mm}
\paragraph{Wan-Animate.}
Wan-Animate uses VitPose~\cite{xu2022vitpose} to extract the facial skeleton for the character in portrait video as head pose representations. Then, it adopts an encoder structure identical to that of LIA~\cite{wang2022latent} to extract expression features from driving frame. Therefore, we combine the facial skeleton of source frame with expression features from driving frame to animate the source frame.

\begin{table*}[t]
  \centering
  \caption{Quantitative comparisons of self-reenactment portrait animation on MEAD~\cite{kaisiyuan2020mead} dataset. The top of the table shows the results of non-diffusion-based methods, while the bottom presents diffusion-based methods.}
  \resizebox{\linewidth}{!}{
  \begin{tabular}{lcccccccccc}
      \toprule[1.5pt]
      Method & PSNR$\uparrow$ & SSIM$\uparrow$ & LPIPS$\downarrow$ & $\mathcal{L}_1$$\downarrow$ & CSIM$\uparrow$ & MAE(\textdegree)$\downarrow$ &AED$\downarrow$ &APD$\downarrow$ &FID$\downarrow$ &FVD$\downarrow$ \\

      \midrule[1pt]
       
       GAGAvatar~\cite{chu2024gagavatar}  &- & - & -& -& 0.8946 & 5.1074 & 0.3781 & 0.0101 & - & - \\
       
       Portrait4D-v2~\cite{deng2024portrait4dv2}  & 20.0907 & 0.7746 & 0.3358 & 0.0617 & 0.8793 & 5.4329 & 0.4353 & 0.0149 & 85.3589 & 460.7595 \\
       
       PD-FGC~\cite{wang2022pdfgc}  & 20.8419 & 0.7824 & 0.341 & 0.0573 & 0.3256 & 8.3772 & 0.7156 & 0.0202 & 92.8886 & 1276.3855 \\
       
       EMOPortrait~\cite{drobyshev2024emoportraits}   & 26.1748 & 0.8729 & 0.1544 & 0.0287 & 0.5959 & 6.6994 & 0.4992 & 0.0128 & 37.5216 & 443.7428 \\

       EDTalk~\cite{tan2024edtalk} & 26.9246 & 0.8964 & 0.1443 & 0.0319 & 0.8592 & 6.218 & 0.4333 & 0.0077 & 43.4199 & 343.9973  \\

       LIA-X~\cite{wang2025lia} & 22.6439 & 0.8232 & 0.1816 & 0.0386 & 0.8957 & 5.3919 & 0.4633 & 0.0636 & 32.4902 & 323.2831  \\

       LivePortrait~\cite{guo2024liveportrait}  & \underline{32.9063} & \underline{0.9464} & \underline{0.0527} & \underline{0.0148} & \underline{0.9379} & \underline{3.5497} & \underline{0.2471} & \underline{0.0041} & \underline{10.4759} & \underline{84.3131} \\
       
       \midrule[1pt]

       FYE~\cite{ma2024follow}  & 27.1819 & 0.8963 & 0.1039 & 0.0243 & 0.8767 & 5.6658 & 0.527 & 0.0109 & 30.5002 & 350.4705 \\
       
       AniPortrait~\cite{wei2024aniportrait}  &29.0281 & 0.9125 & 0.081 & 0.0198 & 0.8904 & 4.8224 & 0.3989 & 0.0077 & 19.9857 & 191.8255 \\
       
       X-NeMo~\cite{zhao2025x}  &22.4136 & 0.7313 & 0.1916 & 0.0551 & 0.8594 & 10.4812 & 0.4168 & 0.0097 & 50.6639 & 409.2123 \\
       
       ReliPA~\cite{guo2025high}  & 24.0052 & 0.8525 & 0.1601 & 0.0409 & 0.8212 & 6.3512 & 0.5174 & 0.0117 & 35.1439 & 455.0235 \\
       
       SkyReels-A1~\cite{qiu2025skyreels}  &25.9931 & 0.8825 & 0.1182 & 0.032 & 0.8668 & 5.7577 & 0.594 & 0.0105 & 22.6852 & 278.6554 \\
       
       Hunyuan-Portrait~\cite{xu2025hunyuanportrait}  & 26.4138 & 0.8779 & 0.0941 & 0.0309 & 0.922 & 4.7961 & 0.3348 & 0.0101 & 18.896 & 139.2772 \\
       
       FantasyPortrait~\cite{wang2025fantasyportrait} & 22.6155 & 0.7789 & 0.1498 & 0.0634 & 0.8622 & 6.606 & 0.5147 & 0.0116 & 35.7586 & 258.8542  \\
       
       Wan-Animate~\cite{cheng2025wan} &  21.9017 & 0.8105 & 0.2159 & 0.054 & 0.827 & 5.7592 & 0.5307 & 0.0144 & 24.9683 & 465.8136\\
       
       VACE~\cite{vace} & 15.0009 & 0.5046 & 0.4083 & 0.1587 & 0.5472 & 10.0836 & 0.745 & 0.021 & 118.5134 & 870.7917\\
       
       AvatarArtist~\cite{liu2025avatarartist} & 18.7405 & 0.7173 & 0.3891 & 0.0774 & 0.7402 & 6.5339 & 0.5571 & 0.0194 & 83.5354 & 815.3565\\ 
       \midrule[1pt]
      \textbf{Ours} & \textbf{33.7235} & \textbf{0.9501} & \textbf{0.0491} & \textbf{0.0125} & \textbf{0.9521} & \textbf{3.1576} & \textbf{0.1971} & \textbf{0.0038} & \textbf{10.132} & \textbf{71.58}   \\
      \bottomrule[1.5pt]
  \end{tabular}
  }
  \label{tab:self-driven}
\end{table*}

\section{More Results} \label{sec:results}
We provide more generated results of our PerformRecast in this section.

\begin{figure*}[t!]
  \centering
  \includegraphics[width=\linewidth]{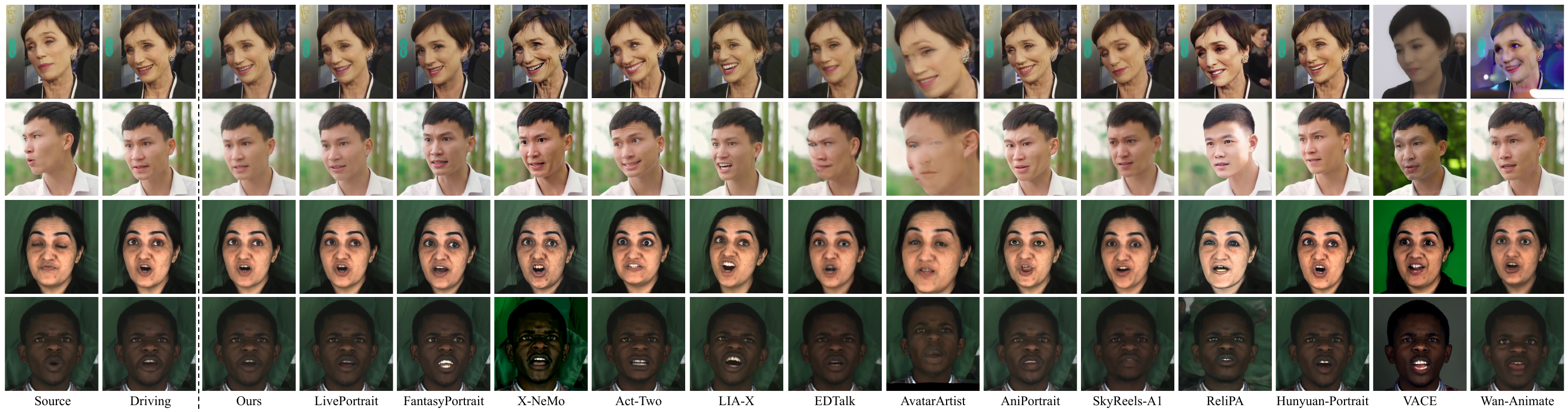}
  \caption{More generated results on self-reenactment task of different methods. The first two source-driving paired images are from VFHQ dataset~\cite{xie2022vfhq} and the last two source-driving paired images are from MEAD dataset~\cite{kaisiyuan2020mead}.}
  \label{fig:self-id-suppl}
\end{figure*}

\begin{figure*}[t!]
  \centering
  \includegraphics[width=\linewidth]{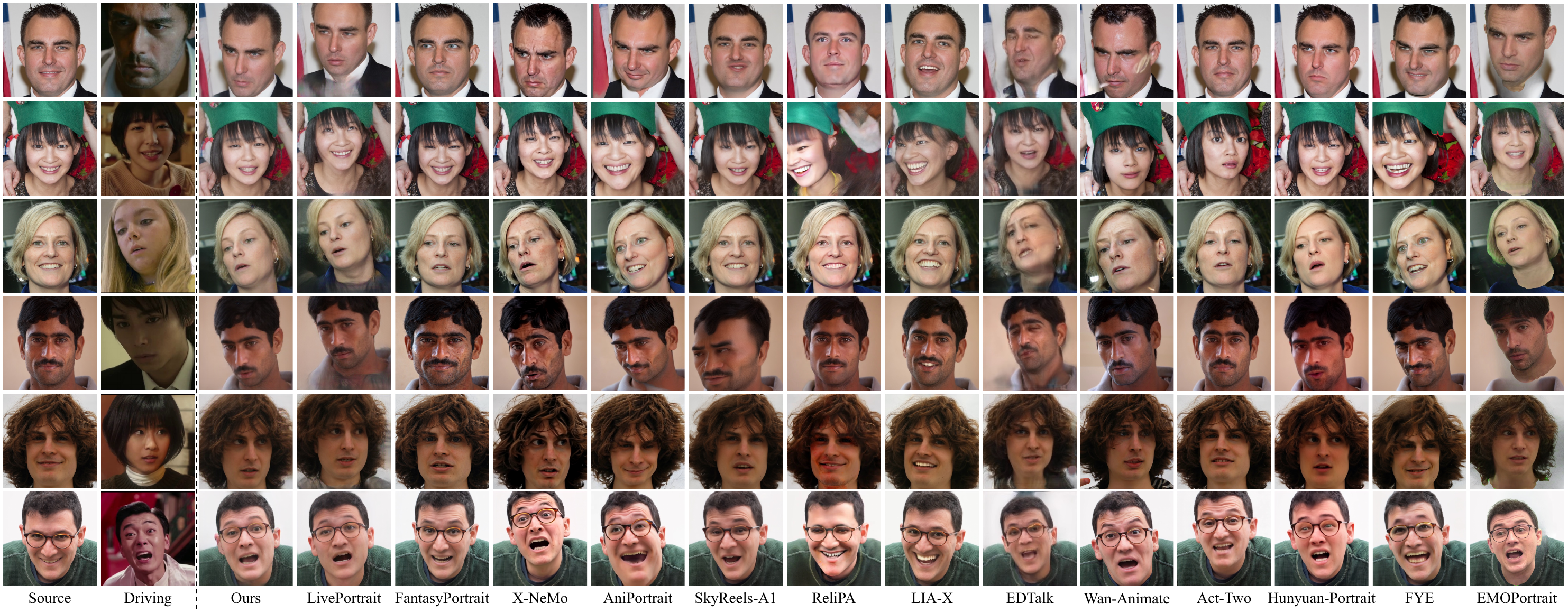}
  \caption{More generated results on cross-reenactment task of different methods. The source images are from FFHQ dataset and the driving frames are from famous films and television clips.}
  \label{fig:cross-id-suppl}
\end{figure*}

\subsection{Portrait Video Expression Editing}
We first compensate for the missing visual comparisons on the enhancement mode in~\cref{fig:enhancement} as mentioned in the main manuscript. LivePortrait~\cite{guo2024liveportrait} generates inaccurate lip movements on the enhancement mode. Act-Two~\cite{runway} tends to synthesize exaggerated mouth movements, leading to less realistic facial animations. On the contrary, our method succeeds in enhancing the facial expressions via adding the expressions of driving video on the top of that of source video.

We then show qualitative results of all the compared methods on our proposed test benchmark in~\cref{fig:expression}. The four modified diffusion-based methods perform extremely poorly on portrait video expression editing task. They all produce incorrect facial expressions with severe artifacts and distortions. As a result, our carefully designed PerformRecast achieves the best performance on both replacement and enhancement modes compared to all previous approaches.

\subsection{Self-reenactment}
We also report the quantitative results of self-reenactment portrait animation on MEAD dataset~\cite{kaisiyuan2020mead} in~\cref{tab:self-driven}. All the methods are evaluated on a random split of MEAD dataset, which consists of 70 videos. As shown in~\cref{tab:self-driven}, our method achieves the best performance across all metrics on MEAD dataset~\cite{kaisiyuan2020mead}, highlighting its superiority over other existing approaches.

What's more, we also present more qualitative results of different compared methods in~\cref{fig:self-id-suppl}. LivePortrait~\cite{guo2024liveportrait} tends to generate blurred results around the eyes in the first and third cases. It also struggles to preserve the subtle expressions in the second and fourth cases. Other diffusion-based methods are prone to generating unstable results and exaggerated facial expressions. On the contrary, our PerformRecast faithfully recovers the driving frames with fine-grained details.

\subsection{Cross-reenactment}
We provide more cross-reenactment portrait animation results generated by our PerformRecast and some other methods in~\cref{fig:cross-id-suppl}. The source images are from FFHQ dataset~\cite{karras2019style} and we use some famous films and television clips as driving frames. From which we can conclude that our method is capable of preserving the head pose, facial expressions and eyeball directions in the driving frames with high fidelity, while generating clear and high-quality images. Although our method does not achieve best performance on all evaluation metrics as reported in the main manuscript, it markedly outperforms all other methods in visual effects. This is most likely because our used quantitative evaluation metrics mainly rely on some pre-trained networks, whose inherent priors may limit their ability to faithfully reflect the actual performance of each method in some scenarios. Developing more evaluation metrics which are capable of accurately assessing the accuracy of head pose, facial expressions and gaze direction is an interesting research direction in the future.

\begin{figure}[t!]
  \centering
  \includegraphics[width=\columnwidth]{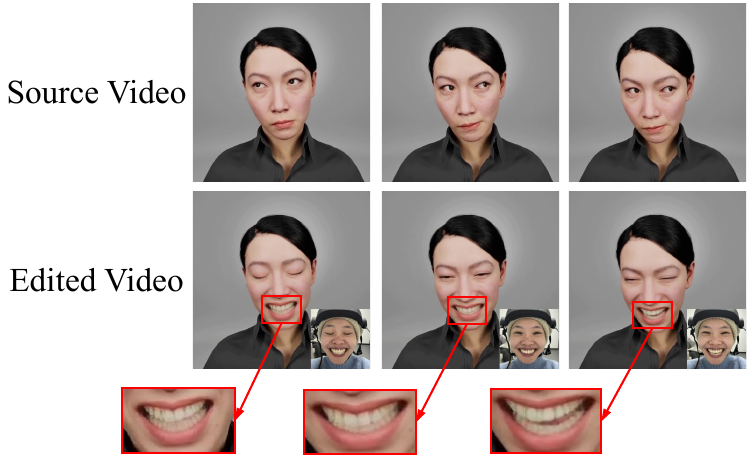}
  \caption{A typical failure case when our method generating teeth while the mouth is closed in source video.}
  \label{fig:limitation}
\end{figure}

\section{Limitations and Discussions} \label{sec:limitation}
In portrait video expression editing task, our method tends to produce blurry results in regions that are not visible in the source video, especially when generating teeth while the mouth is closed in source video. ~\cref{fig:limitation} presents a typical failure case of this scenario. This is mainly because our model is GAN-based, and unlike diffusion-based models, it has limited ability to imagine and synthesize unseen objects. In the future, we are planning to combine the disentangling capability of 3D Morphable Face Model with the generative power of large-scale pre-trained image diffusion models or video diffusion models, aiming to further improve the fidelity and clarity of synthesized videos.

\section{Ethics Statement} \label{sec:ethics}
This work advances portrait animation and portrait video facial expression editing for virtual avatars. Our methods are not intended for malicious use, and all synthesized content should clearly indicate its artificial nature. We acknowledge potential misuse, such as deepfakes, and are developing tools to help detect synthetic videos. At the same time, our technology can support education, communication assistance, and therapeutic applications, reflecting our commitment to responsible and ethical AI development.


\end{document}